\documentclass{article} % For LaTeX2e
\usepackage{colm2024_conference}

\usepackage{microtype}
\usepackage{hyperref}
\usepackage{url}
\usepackage{booktabs}
\definecolor{darkblue}{rgb}{0, 0, 0.5}
\hypersetup{colorlinks=true, citecolor=darkblue, linkcolor=darkblue, urlcolor=darkblue}

\usepackage{hyperref}
\usepackage{url}
\usepackage{xspace}
\usepackage{booktabs}
\usepackage{wrapfig}
\usepackage{float}
\usepackage{natbib}
\usepackage{graphicx}
\usepackage{xcolor}
\usepackage{amsmath}
\usepackage{amsfonts}
\usepackage{algorithm}
\usepackage{algorithmicx}
\usepackage{algpseudocode}
\usepackage{spverbatim}
\usepackage{hyperref}
\usepackage{soul}
\usepackage{tablefootnote}
\usepackage{multirow}
\usepackage{colortbl}
\usepackage{yfonts}
\usepackage{amssymb}
\usepackage{wasysym}
\usepackage{longtable}

\usepackage[T1]{fontenc}
\usepackage{calligra}
\input Zallman.fd

\usepackage[T1]{fontenc}

\usepackage{tabularx} % For tabularx environment
\usepackage{fontawesome}

\usepackage{booktabs} % For nicer tables
\usepackage{array}    % For table wrapping

\definecolor{darkgreen}{rgb}{0.0, 0.5, 0.13} 
\definecolor{gold}{rgb}{0.83, 0.69, 0.52}

\newcommand{\github}{\raisebox{-1.5pt}{\includegraphics[height=1.05em]{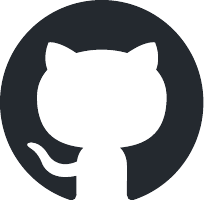}}\xspace}

\newcommand{\claude}{\textsc{\small Claude-3-Opus}}
\newcommand{\chatgpt}{\textsc{\small GPT-3.5-Turbo}}
\newcommand{\gpt}{\textsc{\small GPT-4}}
\newcommand{\gptturbo}{\textsc{\small GPT-4-Turbo}}
\newcommand{\mixtral}{\textsc{\small Mixtral}}

\newcommand{\Faithful}{\textcolor{teal}{\textit{Faithful}}}
\newcommand{\Unfaithful}{\textcolor{purple}{\textit{Unfaithful}}}
\newcommand{\PartialSupport}{\textcolor{violet}{\textit{Partial Support}}}

\newcommand{\fables}{\textsc{\small Fables}}

\title{\raisebox{-0.5ex}{\includegraphics[height=1.4em]{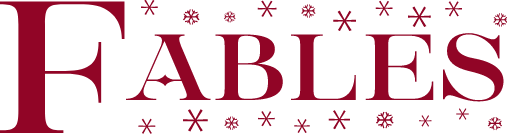}} : Evaluating faithfulness and content selection \\in book-length summarization}

\author{Yekyung Kim\textsuperscript{\faMoonO}, Yapei Chang\textsuperscript{\faMoonO}, Marzena Karpinska\textsuperscript{\faMoonO}, Aparna Garimella\textsuperscript{\faPagelines},\\ \textbf{Varun Manjunatha\textsuperscript{\faPagelines}, Kyle Lo\textsuperscript{\faStarO}, Tanya Goyal\textsuperscript{\faMagic}, Mohit Iyyer\textsuperscript{\faMoonO}}\\
UMass Amherst\textsuperscript{\faMoonO}, Adobe\textsuperscript{\faPagelines}, Allen Institute for AI\textsuperscript{\faStarO}, Princeton\textsuperscript{\faMagic}\\
\texttt{\{yekyungkim, yapeichang, mkarpinska, miyyer\}@umass.edu}\\
\texttt{\{garimell, vmanjuna\}@adobe.com, kylel@allenai.org, tanyagoyal@princeton.edu}} 

\colmfinalcopy
\begin{document}

\maketitle

\begin{abstract}
While long-context large language models (LLMs) can technically summarize book-length documents ($>100$K tokens), the length and complexity of the documents have so far prohibited evaluations of input-dependent aspects like faithfulness. In this paper, we conduct the first large-scale human evaluation of faithfulness and content selection on LLM-generated summaries of fictional books. Our study mitigates the issue of data contamination by focusing on summaries of books published in 2023 or 2024, and we hire annotators who have fully read each book prior to the annotation task to minimize cost and cognitive burden. 
We collect \fables, a dataset of annotations on 3,158 claims made in LLM-generated summaries of 26 books, at a cost of \$5.2K USD, which allows us to rank LLM summarizers based on faithfulness: \claude\ significantly outperforms all closed-source LLMs, while the open-source \mixtral\ is on par with \chatgpt. An analysis of the annotations reveals that most unfaithful claims relate to events and character states, and they generally require indirect reasoning over the narrative to invalidate.  
While LLM-based auto-raters have proven reliable for factuality and coherence in other settings, we implement several LLM raters of faithfulness and find that none correlates strongly with human annotations, especially with regard to detecting unfaithful claims. Our experiments suggest that detecting unfaithful claims is an important future direction not only for summarization evaluation but also as a testbed for long-context understanding. Finally, we move beyond faithfulness by exploring content selection errors in book-length summarization: we develop a typology of omission errors related to crucial narrative elements and also identify a systematic over-emphasis on  events occurring towards the end of the book.
We release \fables\ to spur further research on the evaluation of book-length summarization.

\begin{center}
    \renewcommand{\arraystretch}{1.2}
    \begin{tabular}{rl}
         \github & \href{https://github.com/mungg/FABLES}{\path{https://github.com/mungg/FABLES}} \\
    \end{tabular}
\end{center}

\end{abstract}
\section{Introduction}\label{sec:intro}
Advances in long-context language models have sparked interest in summarizing book-length documents ($>$100K tokens). 
Despite the importance of faithfulness and content relevance for summary quality, recent work in this regime  focuses only on input-agnostic aspects like coherence \citep{Chang2023BooookScoreAS}. 
This is due to the length and complexity of the input documents: hiring human annotators to read and understand them is expensive and time-consuming. 
Our work fills this gap by presenting the first large-scale human evaluation of faithfulness and other content selection errors in book-length summarization.

We mitigate challenges associated with document complexity by hiring workers who have already read a book published in 2023 or 2024 (to avoid data contamination) for enjoyment prior to beginning the annotation task. We produce summaries for these books via five configurations of the hierarchical summarization methodology described in~\citet{Chang2023BooookScoreAS}, each of which varies the base LLM and chunk size. Following prior work on faithfulness and factuality evaluation, such as LongEval~\citep{krishna-etal-2023-longeval} and FactScore~\citep{DBLP:conf/emnlp/MinKLLYKIZH23}, we decompose each summary into a list of claims which are then individually verified against the input document.

\begin{figure}[tbp]
\centering
\includegraphics[width=0.95\linewidth]{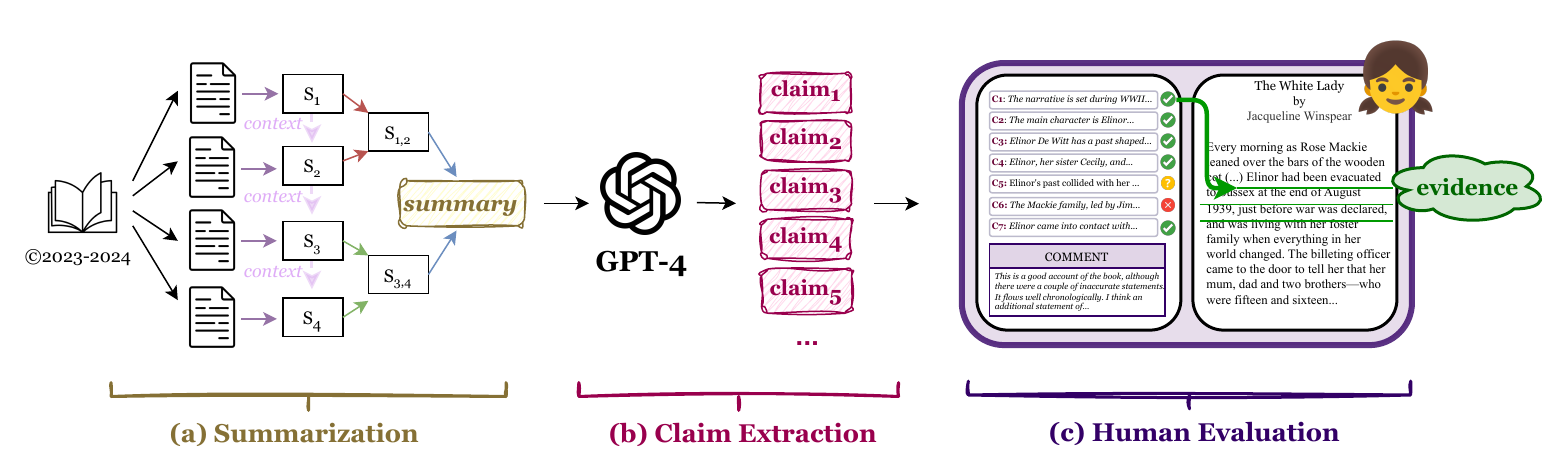}
\caption{Our pipeline for collecting faithfulness annotations in book-length summarization (\fables). First, \textbf{(a)} we generate summaries through hierarchical merging. Next, \textbf{(b)} we prompt \gpt\ to extract decontextualized claims. Finally, \textbf{(c)} we conduct a human evaluation of these claims, requiring annotators to validate each claim and provide their reasoning and evidence for the assigned label.}
\label{fig:pipeline}
\end{figure}

In total, our \fables\ dataset (\textbf{F}aithfulness \textbf{A}nnotations for \textbf{B}ook-\textbf{Le}ngth \textbf{S}ummarization) contains \textbf{3,158} claim-level annotations of faithfulness across \textbf{26} narrative texts, along with evidence for each claim in the form of quotations from the book as well as free-form comments at both the claim and summary level (Figure~\ref{fig:pipeline}).\footnote{While we cannot release the book text due to copyright restrictions, we publicly release all summaries and annotations.} Overall, we observe that \claude\ is the most faithful book-length summarizer by a significant margin, followed by \gptturbo. Beyond ranking LLMs, our annotations also shed light on the following previously unexplored questions:

\vspace{-0.2cm}
\paragraph{What kinds of faithfulness errors do LLM summarizers make? (\S\ref{sec:annotation})} A qualitative analysis of \fables\ reveals that the majority of claims marked as unfaithful are related to \emph{events} or \emph{states} of characters and relationships. Furthermore, most of these claims can only be invalidated via multi-hop reasoning over the evidence, highlighting the task`s complexity and its difference from existing fact-verification settings \citep{DBLP:conf/emnlp/MinKLLYKIZH23, kamoi2023wice}.

\vspace{-0.2cm}
\paragraph{Can faithfulness be evaluated automatically? (\S\ref{sec:exp})}
Collecting human annotations on 26 books cost us \textbf{\$5.2K}, demonstrating the difficulty of scaling our workflow to new domains and datasets. We thus implement multiple LLM-based raters of faithfulness, following prior work such as BooookScore~\citep{Chang2023BooookScoreAS} and FactScore~\citep{DBLP:conf/emnlp/MinKLLYKIZH23} that achieve high correlation with human judgments. However, all of our metric configurations struggle to reliably identify unfaithful claims. Our best-performing method operates similarly to ``needle-in-the-haystack''-style evaluations \citep{Needle_In_A_Haystack, reid2024gemini} by feeding as much of the book as possible into a long-context LLM along with a single claim to verify. We promote this claim-level verification task as both important for book-length summarization evaluation as well as a challenging benchmark for long-context understanding.

\vspace{-0.2cm}
\paragraph{What other errors, beyond faithfulness, do LLM summarizers make? (\S\ref{sec:other-errors})} By coding all of the summary-level free-form comments in \fables, we find that annotators frequently point out \emph{omissions} of critical information. We develop the first taxonomy of omission errors in book-length summarization and observe that key events, details, and themes are frequently omitted by all LLMs. 
We also observe other content selection errors: for example, even our strongest summarizers, \claude\ and \gptturbo, over-emphasize content towards the end of books to the detriment of the beginning.

All prompts used in this paper can be found in \S\ref{app:prompt-section}.

\section{Collecting human annotations}
\label{sec:data_method}
In this section, we describe our 
pipeline for collecting \fables, which consists of human annotations of both faithfulness and overall quality of LLM-generated book summaries.

\paragraph{Collecting a corpus of newly-published fictional books:}

It is infeasible, both in terms of cost and time, to ask annotators to read long  books ($\geq100K$ tokens) for the sole purpose of annotating LLM-generated summaries. While we can remove this burden by choosing famous books that many people have already read, such as those in BookSum \citep{kryscinski-etal-2022-booksum}, LLMs have also likely seen these books and their summaries during pretraining~\citep{chang-etal-2023-speak}, which can skew the evaluation of generated claims. Instead, we use an annotator-driven workflow to sidestep these issues. We recruit a pool of annotators via Upwork\footnote{\url{https://www.upwork.com}} who self-report having read one or more English books published in 2023 or 2024. Our final annotator pool consists of \textbf{14} native English speakers,  and we purchase electronic copies of 26 books listed by them.\footnote{We convert epubs to text files preserving all information including front and back matter.} The mean length of books in our dataset is \textbf{121K} tokens (see \autoref{tab:data_stats} for statistics).

\begin{table}[t]
\centering
\footnotesize
\begin{tabular}{lcccccc}
\hline
\addlinespace
& \multicolumn{3}{c}{\textbf{Books } \includegraphics[height=1.1em]{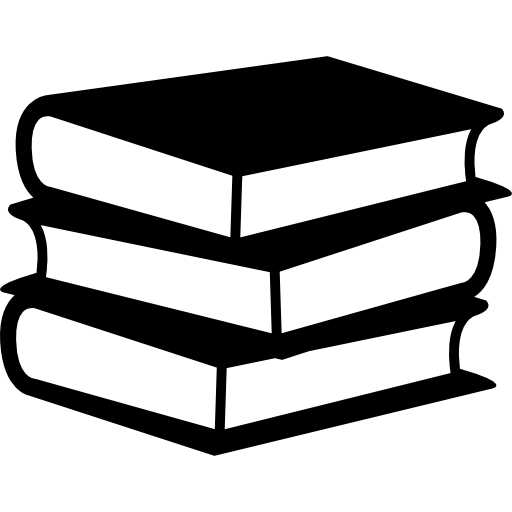}} & \multicolumn{3}{c}{\textbf{Annotations } \includegraphics[height=1em]{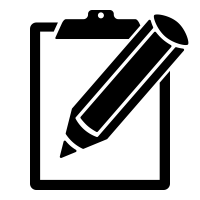}} \\
\cmidrule(r){2-4} \cmidrule(lr){5-7}
 & {\renewcommand{\arraystretch}{1}\begin{tabular}[c]{@{}c@{}}\textbf{ Documents }\\\textit{(n=26) }\vspace{.0em}\end{tabular}} & {\renewcommand{\arraystretch}{1}\begin{tabular}[c]{@{}c@{}}\textbf{ Summaries }\\\textit{(n=130) }\vspace{.0em}\end{tabular}} & {\renewcommand{\arraystretch}{1}\begin{tabular}[c]{@{}c@{}}\textbf{ Claims }\\\textit{(n=3,158) }\vspace{.0em}\end{tabular}} & {\renewcommand{\arraystretch}{1}\begin{tabular}[c]{@{}c@{}}\textbf{ Reasons }\\\textit{(n=1,513) }\vspace{.0em}\end{tabular}} & {\renewcommand{\arraystretch}{1}\begin{tabular}[c]{@{}c@{}}\textbf{ Evidence }\\\textit{(n=3,051) }\vspace{.0em}\end{tabular}} & {\renewcommand{\arraystretch}{1}\begin{tabular}[c]{@{}c@{}}\textbf{ Comments }\\\textit{(n=130) }\vspace{.0em}\end{tabular}}\\ 
\hline
\addlinespace
Mean     & 121,467      & 594.3  & 19.8 & 37.6 & 194.7 &  155  \\
St. dev. & 35,836    & 119.5    & 6.4 & 33.4 & 218.5 &  148.4 \\
Max         & 243,965   & 798   & 58 & 281 & 2435 &  823  \\
Min         & 49,762 & 172   & 6 & 2 & 5 &  6 \\ \hline
\end{tabular}
\caption{Number of tokens across books and \fables\ annotations; based on \texttt{tiktoken} (\url{https://github.com/openai/tiktoken}) tokenizer.}
\label{tab:data_stats}
\end{table}

\paragraph{Prompting LLMs to generate book summaries:} To summarize book-length documents, we adopt the hierarchical merging strategy from ~\citep{Chang2023BooookScoreAS}; see Figure~\ref{fig:pipeline} for an illustration of the method. We use \chatgpt, \gpt, \gptturbo\ \citep{DBLP:journals/corr/abs-2303-08774}, \mixtral\ \citep{jiang2024mixtral}, and \claude\ \citep{Anthropic2023Claude} as the backbone models.\footnote{All summaries were generated in February 2024 using the following checkpoints: \texttt{gpt-3.5-turbo}, \texttt{gpt-4-0613}, \texttt{gpt-4-0125-preview}, \texttt{Mixtral-8x7B-Instruct-v0.1}, and \texttt{claude-3-opus-20240229}.}

\begin{figure}[t]
\centering
\includegraphics[width=1\linewidth]{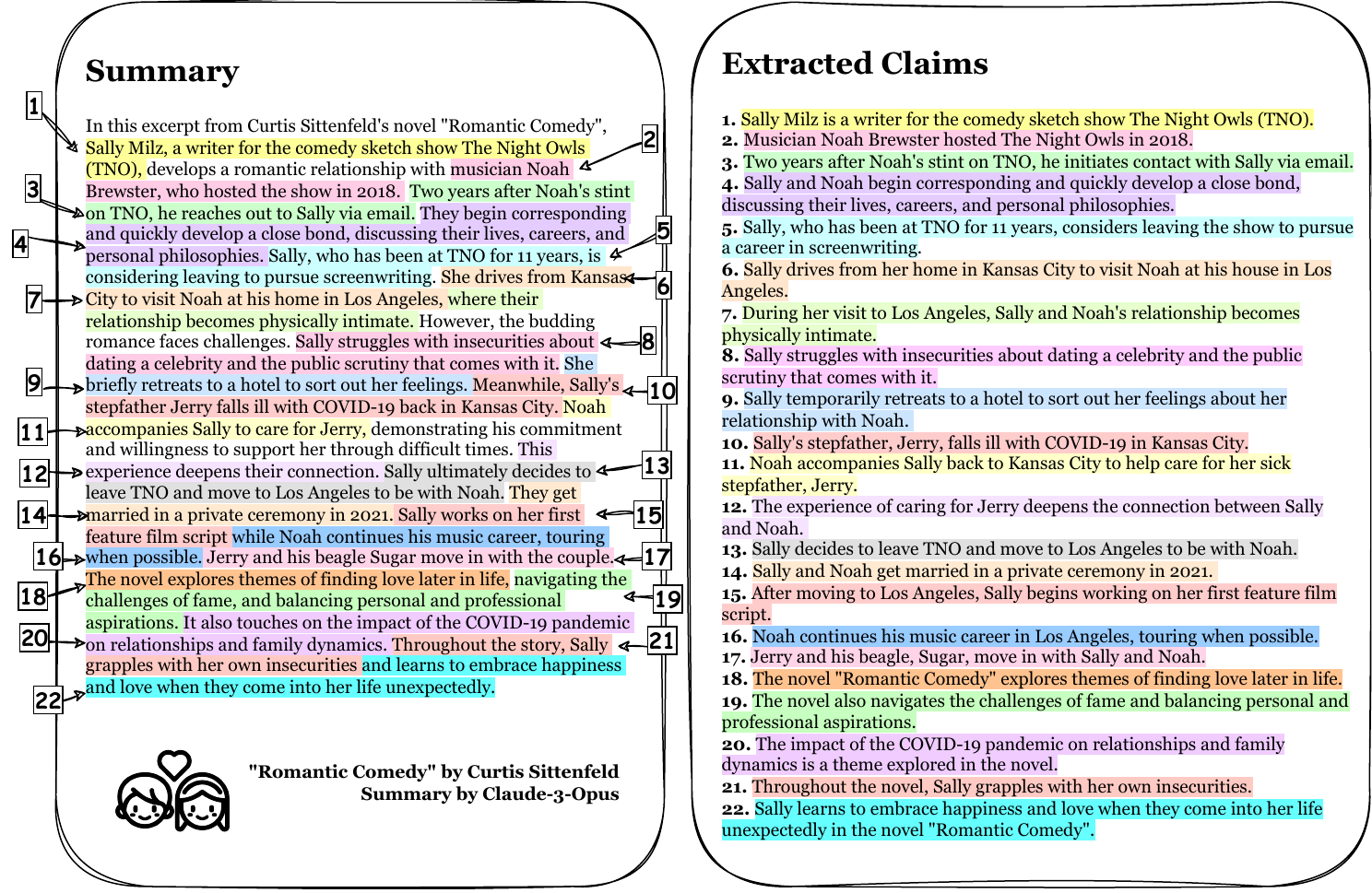}
\caption{Example summary generated by \claude\ and claims extracted by \gpt.}
\label{fig:claim_extraction_example}
\end{figure}

\paragraph{Decomposing summaries into claims:}
Following prior works on evaluating long-form summary faithfulness~\citep{krishna-etal-2023-longeval,DBLP:conf/emnlp/MinKLLYKIZH23,wei2024longform}, we decompose our summaries into \emph{atomic claims} to enable fine-grained annotation. We prompt an LLM (\gpt) with two primary instructions: (1) each atomic claim must be fully understandable on its own without requiring additional context from the summary (e.g., resolved pronouns), and (2) whenever possible, each claim should be situated within its relevant temporal, locational, and causal context.
Human validation by the authors of a random sample of 100 extracted claims demonstrated 100\% precision (i.e., each claim can be traced to the summary without any extra or incorrect information). See \autoref{fig:claim_extraction_example} for example of summary and its extracted claims; see \S\ref{app:prompt-section} for exact prompt and \S\ref{app:recall_decomp} for recall analysis.

\paragraph{Collecting human annotations:} 
The Upwork annotators were tasked with two primary objectives: 
\begin{itemize}
    \item \textbf{Claim-level:} Assess the faithfulness of claims extracted from model-generated summaries of their assigned book(s). Annotators reviewed claims made about their selected book(s) and determined their accuracy by choosing one of four options for each decomposed claim: (a) \Faithful\ -- accurate reflection of the narrative, (b) \Unfaithful\ -- misrepresentation of the narrative, (c) \PartialSupport\ -- partially corroborated by the narrative, or (d) \textit{Can't verify} -- indeterminable. They provided free-form textual justifications to support their selections, including \textit{evidence} in the form of quotations from the book when relevant.
    \item \textbf{Summary-level:} Provide free-form summary-level comments on the overall quality of the summaries. Annotators critiqued the claim set as a whole, identifying omissions, inaccuracies, disproportionate emphasis on trivial plot points, or other concerns. 
\end{itemize}

\noindent The annotators used a customized interface,\footnote{Refer to \S\ref{app:hum-eval-section} for the screenshots of the interface and the exact wording of the task.} which provided them full access to the book text for reference. 
Each annotator was assigned to annotate all five LLM-generated summaries for their assigned book, which were presented in a randomized order. Annotators received \$200 for this task, which took $\sim$11 hours to complete (STD=6.34). In total, \fables\ contains 3,158 annotated claims from 130 summaries across 26 books at a cost of \$5.2K USD. We assess the quality of our dataset using inter-annotator agreement and self-consistency metrics. More details can be found in \S\ref{app:quality_annotation}.

\section{Developing a taxonomy of faithfulness errors in \textsc{Fables}}\label{sec:annotation}

In this section, we present results from our statistical and qualitative analysis of the 3,158 claim-level faithfulness annotations in \fables, which include both free-form comments and citation evidence to support or refute these claims.\footnote{For 107 claims, the annotators were unable to cite evidence either in favor or against the claim.} Broadly, we observe that \claude\ is the most faithful LLM summarizer, with 90\% of its claims rated as faithful, followed by \gpt\ and \gptturbo\ at 78\%, \chatgpt\ at 72\%, and \mixtral\ at 70\% (\autoref{tab:human_annotations_results}).

\begin{table}[htbp]
    \centering
    \scalebox{0.75}{
    \begin{tabular}{lcc|cccc}
    \toprule
    \textbf{Model}          &  \textbf{Chunk size} & \textbf{Avg \# Claims$_{{\textbf{STD}}}$} & \textbf{Faithful} & \textbf{Unfaithful} & \textbf{Partial support} & \textbf{Can't verify} \\
    \midrule
    \chatgpt       & 2,048     
    & 23.23$_{{3.29}}$       
    & 71.52 & 11.26   & 13.08        & 4.14      \\
    \mixtral        & 2,048      & 27.50$_{{4.61}}$         & 68.67 & 11.47   & 17.2        & 2.66      \\
    \gpt\tablefootnote{Note that \gptturbo, \mixtral, and \gpt\ are capable of processing sequences of up to 16K, 32K, and 8K tokens respectively. In this study we opted to use a shorter context window to maintain consistency with the experimental setup described in \cite{Chang2023BooookScoreAS}.}            & 2,048      & 26.23$_{{3.54}}$             & 78.15 & 4.55    & 15.98        & 1.32      \\
    \gptturbo    & 100,000    & 21.65$_{{2.35}}$         & 77.62 & 7.64   & 12.08        & 2.66      \\
    \claude  & 180,000    & 22.85$_{{4.87}}$         & 90.89 & 2.1    & 6.65         & 0.35      \\
    \bottomrule
\end{tabular}
    }
    \caption{
    Percentage of claims extracted from LLM-generated summaries rated by humans as \textit{faithful}, \textit{unfaithful}, \textit{partial support} or \textit{can't verify}.
    Chunk size denotes the token count per chunk used for summarization across models; we also include the mean and standard deviation of claim counts in generated summaries. Please note that the percentage of each label for \claude\ is calculated from 24 out of 26 books. The model was unable to merge summaries for two books due to content discrepancies. 
    }
    \label{tab:human_annotations_results}
\end{table}

\paragraph{Analysis of unfaithful claims:} To further study the nature of unfaithful claims, we characterize all 205 such claims along two dimensions: \textsc{claim type} and \textsc{reasoning type} (see \autoref{tab:faith-anno-results} for taxonomy and frequency counts).\footnote{There are actually 247 annotations with unfaithful claims, but for this analysis we leave out 42 unclear ones that require further clarification from the annotators. Note that since the claims sometimes contain multiple subclaims, we allow each annotation to have multiple labels.} Most unfaithful claims are about specific \textit{events} (31.5\%) or the \textit{state} of some character or relationship (38.6\%). Crucially, a majority of unfaithful claims require \textit{indirect reasoning} to refute (50.2\%), making this a more challenging faithfulness evaluation setting compared to prior work \citep{kamoi2023wice, DBLP:conf/emnlp/MinKLLYKIZH23}. More details on this analysis can be found in \S\ref{app:criteria}.

\begin{table}[ht]
\centering
\footnotesize
\scalebox{0.87}{
\begin{tabular}{p{0.12\textwidth}p{0.06\textwidth}p{0.45\textwidth}p{0.4\textwidth}}
\toprule
\multicolumn{1}{c}{\textsc{Label}} & \multicolumn{1}{c}{\textsc{Freq}} & \multicolumn{1}{c}{\textsc{Example claim}} & \multicolumn{1}{c}{\textsc{Reason for rejection}} \\
\midrule
\addlinespace
\multicolumn{4}{c}{\includegraphics[height=1.1em]{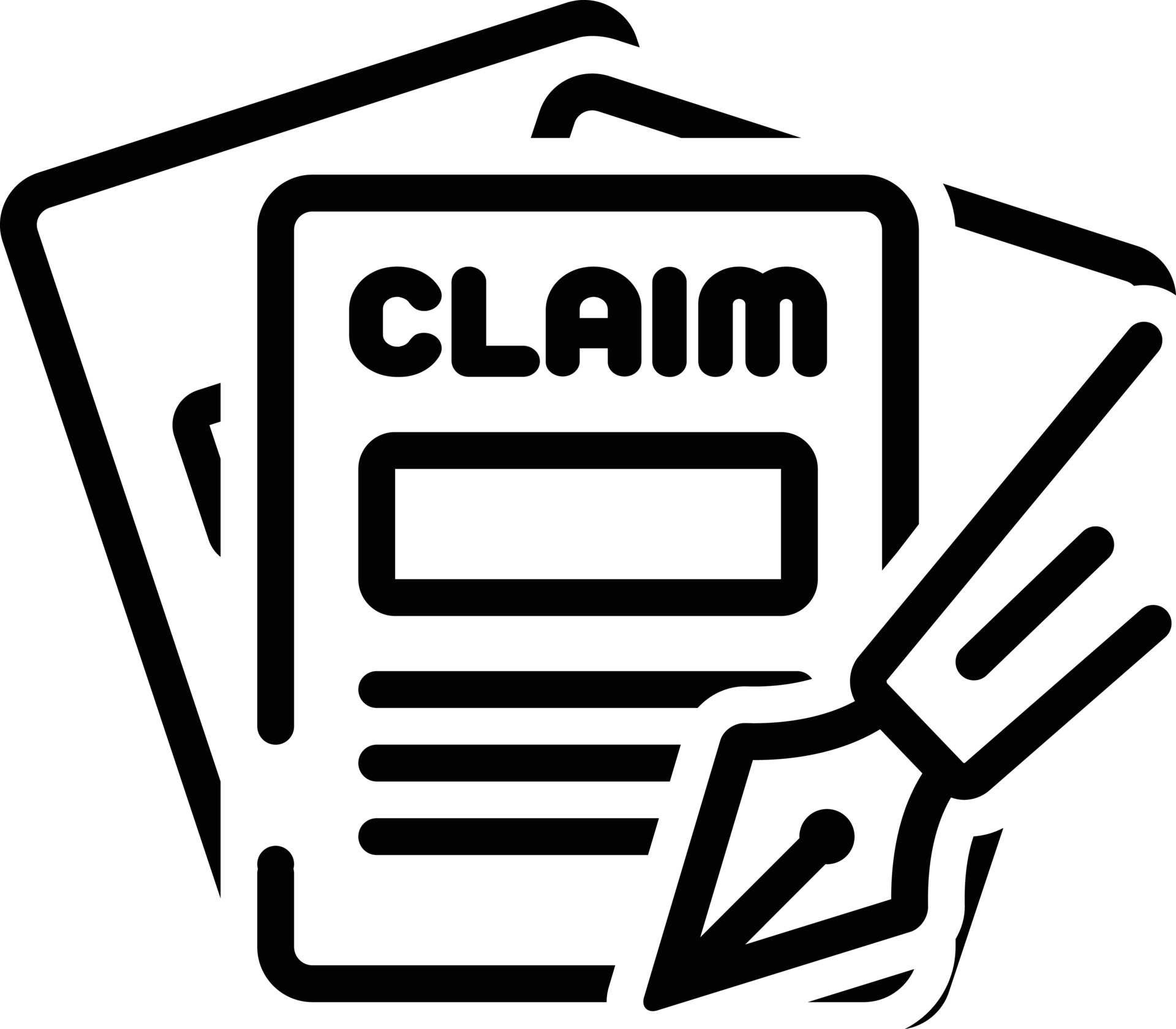}\textbf{ Claim Type}} \\
\addlinespace
\midrule
State & 38.6 & \textit{Roman Kitt is {\color{purple}under pressure} from his father to join the family business.} & Roman is not under pressure, his father bribes people so he gets his dream job. \\
\midrule
Event & 31.5 & \textit{Patricia Liu, Athena’s mother, discovers that June has sold Athena’s manuscript and \textcolor{purple}{confronts} her.} & Patricia never confronts June. \\
\midrule
Cause/effect & 11.2 & \textit{Lilly's abusive ex-boyfriend, Alan Bushy, becomes a suspect \textcolor{purple}{due to the meticulous nature of the murders}.} & He becomes a suspect because he was abusive to Lilly. \\
\midrule
High-level & 11.2 & \textit{The narrative is {\color{purple}non-linear} and features flashbacks, switches between alternate worlds or viewpoints, and present-day conversations between Sally and Danny.} & The narrative is largely linear. \\
\midrule
Introspection & 7.5 & \textit{Juniper Song encounters Athena Liu at a literary event, triggering {\color{purple}feelings of admiration, intimidation, and self-doubt}.} & No part of the book shows that Juniper admires Athena. \\
\midrule
\addlinespace
\multicolumn{4}{c}{\includegraphics[height=1.3em]{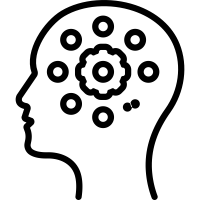}\textbf{ Reasoning Type}} \\
\addlinespace
\midrule
Indirect & 50.2 & \textit{Dean stirs up {\color{purple}tensions} with palace server Fawn.} & This encounter is merely Rennick being protective of Amelia, tension can't be inferred from the book. \\
\midrule
Direct & 36.8 & \textit{The narrative reveals that Maggie had a {\color{purple}brief affair} with a doctor named Danny in Bangkok while she was being followed by unknown entities.} & The book directly states that they are married, so it's not a brief affair. \\
\midrule
Subjective & 7.2 & \textit{Forest is torn between his desire to protect Iris and {\color{purple}confronting his past actions}.} & I don't think Forest makes any real effort to confront his past actions \\
\midrule
Extra info & 5.7 & \textit{The book ``Wildfire" is {\color{purple}the first} in the Icebreaker series.} & It's not stated in the book, but this is actually the second in the series. \\
\bottomrule
\end{tabular}}
\caption{Taxonomy of faithfulness errors with respect to claim type and reasoning type in \fables. For each label, we report its \textit{frequency} and provide an \textit{example} claim-reason pair. More examples and the general labeling scheme can be found in \autoref{tab:faith-anno-criteria}.}
\label{tab:faith-anno-results}
\end{table}

\section{Challenges with automatic faithfulness evaluation}\label{sec:exp}
While insightful, human annotation of faithfulness in book-length summarization is simply not scalable: our annotations cost \$40 USD per summary for a total cost of \$5.2K USD, which is prohibitively expensive for usage during model development and with bigger corpora. In this section, inspired by methods such as FactScore~\citep{DBLP:conf/emnlp/MinKLLYKIZH23} and BooookScore~\citep{Chang2023BooookScoreAS}, we develop LLM-powered automatic raters of faithfulness that operate at the claim level. However, our best method, which relies on prompting \claude\ with the entire book to verify a single claim, is expensive and unreliable at detecting unfaithful claims in \fables, suggesting important directions for future work.

\paragraph{Automatic raters of faithfulness:}
We implement our automatic raters by prompting an LLM in a zero-shot manner to verify a single claim given evidence from the book (\autoref{tab:prompt_templates_claim_eval}), where the evidence can be one of the following:
\begin{itemize}
\item \textbf{None}: As a lower bound, we evaluate the faithfulness of claims without any evidence from the book.

\item \faUsers \textbf{ Human evidence}: We can also use human-annotated evidence from \fables\ obtained via the pipeline described in \S\ref{sec:data_method}. 
This evidence is always related to the claim, but it often takes the form of short, highly-contextual spans that may or may not be sufficient to support claim verification.

\item \faDatabase \textbf{ BM25 retrieval}: We employ BM25 \citep{Robertson1995Okapi} to retrieve passages from the book using the claim as a query. 
We concatenate the $k$ most relevant passages to use as evidence for our evaluation prompt. We set $k = 5$ and chunk passages up to 256 tokens. See \S\ref{app:ablation_study} for performance changes when varying passage length.

\item \faBook \textbf{ Entire book (EB)}: Retrieval is especially challenging in our setting due to the complexity of both the query and document. Intuitively, long-context LLMs can bypass explicit retrieval by simply fitting the entire book into the context as evidence. This setting resembles ``needle-in-the-haystack'' evaluations of prior work~\citep{Needle_In_A_Haystack, levy2024task}, except that it tests a much deeper understanding of the input document.

\end{itemize}

\paragraph{Dataset for experiments:}
Due to budget constraints associated with the ``entire book'' setting, we select seven books, each shorter than 125K tokens, to evaluate the performance of our auto-rater configurations.  This results in 723 total claims, 69 of which are marked as \Unfaithful\ and 654 as \Faithful\ by our human annotators. Note that we do not consider partially supported or unverifiable claims in our experiments due to the increased subjectivity associated with these labels. 
Detailed information regarding this dataset and experiment costs can be found in \S\ref{app:experiment}.

\paragraph{Results:}  We evaluate the performance of each auto-rater configuration by comparing its predictions to the ground-truth labels (\Faithful\ and \Unfaithful) from our human annotations. 
Due to the class imbalance, we report separate F1 scores for each label, split across claims generated by different LLMs, in~\autoref{tab:benchmark_results_hay}.\footnote{We note that scores for \Unfaithful\ claims on a per-model level should be taken with a grain of salt due to the small sample size, particularly for \claude\ summaries.} As a sanity check, the ``no evidence'' setting performs extremely poorly; more interestingly, human evidence underperforms both retrieval and the entire book setting, suggesting that the LLM requires more context to judge claim validity. The best performing auto-rater is \claude\ in the entire book setting, which significantly outperforms both \gptturbo\ in the same setting as well as BM25.

\begin{figure}[htbp]
\centering
\includegraphics[width=0.95\linewidth]{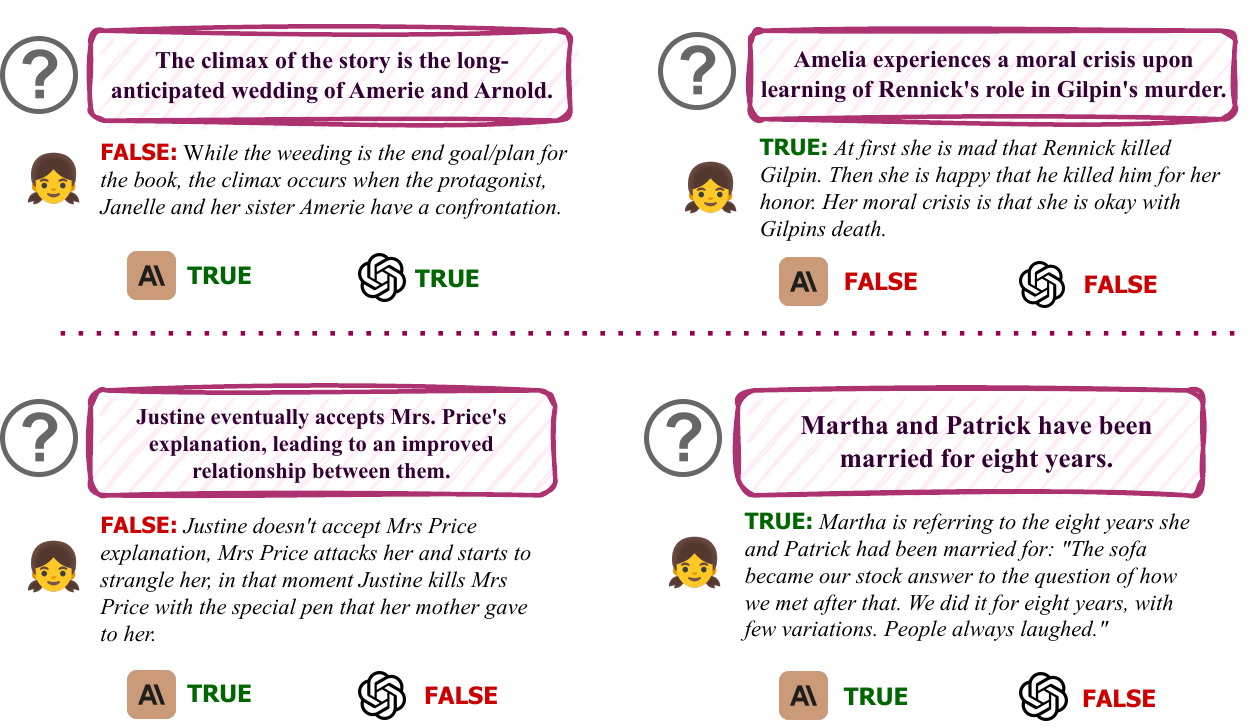}
\caption{Examples of mistakes in label prediction made by \claude\ and \gptturbo\  accompanied by annotator labels and reasoning. More examples can be found in \autoref{fig:examples-nli-results}.}
\label{fig:examples-nli-results-in-main}
\end{figure}

\paragraph{Conclusion:}
Despite it having the best performance in \autoref{tab:benchmark_results_hay}, \claude\ ultimately performs too poorly to be a reliable auto-rater (58.2 F1 when classifying \Unfaithful\ claims).
This comes as a surprise as this pattern of decompose-then-verify has been shown to correlate with human judgments in other settings, like \citet{DBLP:conf/emnlp/MinKLLYKIZH23}.
Manual analysis of the errors
reveals that \claude\ struggles most with
claims involving non-narrative information (23.1\%), assessments often based on common sense reasoning (20.5\%), and character confusions (12.8\%), which often require a deep understanding of the entire book; see confusion matrix in \autoref{fig:examples-nli-results-in-main} and more details in \S\ref{tab:gpt4_turbo_vs_claude3-nli-confusion}.
Qualitatively, we can also gauge from annotator comments (\autoref{tab:annotators-comments-on-task}) the difficulty of this claim verification task as evidence may be difficult to localize (in ``needle-in-the-haystack'' manner) and require full document reasoning.

% tab:annotators-comments-on-task
\begin{table}[h]
\centering
\footnotesize
\begin{tabular}{@{}p{0.03\linewidth}p{0.9\linewidth}@{}}
\toprule
 & \textsc{\faUsers\ Comments} \\ 
\midrule
\addlinespace
\faCommentO & \textit{The hardest part was that some of the claims were very general about the text, {\color{purple}such as describing overall character arcs, which made it hard to find specific textual support.}}  \\
\addlinespace
\faCommentO & \textit{The most difficult part for me was how general some of the sentences were. Because the material was so broad, I felt that I could use 20 or 30 quotations. For example, this book is about many stories of a private investigator in Africa (not exactly what it said, but close). {\color{purple}I could recite the entire book.}} \\
\addlinespace
\faCommentO & \textit{The most difficult part for me was finding supporting quotations for claims that were more abstract (e.g. "The book grapples with the scars of colonialism."). Although I was able to tell right away whether the claim was true or false, based on my own reading, it was at times difficult to find a specific quotation that best proved the claim. {\color{purple}The themes were more often implicit in the text, rather than explicit.}} \\
\addlinespace
\faCommentO & \textit{The most difficult part was to give citations for claims about writing style and intent. The reason was that {\color{purple}these claims are usually based on the book as a whole, so an accurate citation would be the whole book.}} \\ 
\bottomrule
\addlinespace
\end{tabular}
\caption{Annotator comments highlighting the challenges in evidence retrieval.}
\label{tab:annotators-comments-on-task}
\end{table}

\paragraph{Discussion:} It is generally agreed that benchmarking the faithfulness of LLM-generated text is important. However, recent efforts have primarily focused on verifying entity-centric facts \citep{DBLP:conf/emnlp/MinKLLYKIZH23}. Our work, and others \citep{zhu2023annotating, tang2024tofueval, mishra2024fine}, show that these do not provide coverage over all types of LLM errors, especially in more challenging settings like book summarization. Moreover, the retrieve-then-verify framework that forms the backbone of most past evaluation techniques \citep{bohnet2022attributed, gao2023enabling} completely fails for our significantly more challenging setting. Given this evidence, we call for broadening the scope of error types and task settings (including our current task of book-length  summarization) considered by current faithfulness evaluation benchmarks.

\begin{table}[htbp]
\centering
\scalebox{0.63}{
\renewcommand{\arraystretch}{1.2}
\begin{tabular}{l c c c c c c c c c c c}
\toprule
 \multirow{3}{*}{\textit{Summarized by}} & \multicolumn{2}{c}{\textbf{No-Context}} & \multicolumn{2}{c}{\textbf{Human Evidence}} & \multicolumn{2}{c}{\textbf{BM25}} & \multicolumn{2}{c}{\textbf{Entire Book}} & \multicolumn{2}{c}{\textbf{Entire Book}} \\
 
 & \multicolumn{2}{c}{\textbf{\gptturbo}} & \multicolumn{2}{c}{\textbf{\gptturbo}} & \multicolumn{2}{c}{\textbf{\gptturbo}} & \multicolumn{2}{c}{\textbf{\gptturbo}} & \multicolumn{2}{c}{\textbf{\claude}} \\
 
\cmidrule(lr){2-3} \cmidrule(lr){4-5} \cmidrule(lr){6-7} \cmidrule(lr){8-9} \cmidrule(lr){10-11}
 & \textbf{\Faithful}  & \textbf{\Unfaithful}
 & \textbf{\Faithful}  & \textbf{\Unfaithful}    & \textbf{\Faithful}  & \textbf{\Unfaithful}
   & \textbf{\Faithful}  & \textbf{\Unfaithful}    & \textbf{\Faithful}  & \textbf{\Unfaithful}  \\
\midrule
\addlinespace
\chatgpt & 0.396 & 0.248 & 0.686 & 0.369 & 0.801 & 0.373 & 0.887 & 0.357 & 0.929 & 0.619\\
\mixtral & 0.248 & 0.178 & 0.760 & 0.361 & 0.807 & 0.312 & 0.946 & 0.440 & 0.962 & 0.645 \\
\gpt & 0.337 & 0.146 & 0.657 & 0.229 & 0.739 & 0.162 & 0.909 & 0.230 & 0.959 & 0.600 \\
\gptturbo & 0.261 & 0.217 & 0.680 & 0.264 & 0.794 & 0.241 & 0.918 & 0.109 & 0.945 & 0.600 \\
\claude & 0.242 & 0.018 & 0.510 & 0.022 & 0.692 & 0.000 & 0.962 & 0.000 & 0.971 & 0.000 \\
\midrule
Overall & 0.305 & 0.167 & 0.675 & 0.259 & 0.779 & 0.249 & 0.932 & 0.386 & \textbf{0.955} & \textbf{0.582} \\ 
\bottomrule
\end{tabular}
}
\caption{F1 scores for \Faithful\ and \Unfaithful\ label across models with evaluators on 7 books. The best results of each label are in bold. Entire Book refers to the entire book method evaluating faithfulness from large (125k) chunks using either \gptturbo\ or \claude.}
\label{tab:benchmark_results_hay}
\end{table}

\section{Beyond faithfulness: content selection errors in book summarization}\label{sec:other-errors}

As book-length summarization is still a nascent area,  research into other error types beyond coherence~\citep{Chang2023BooookScoreAS} and faithfulness (\S\ref{sec:annotation}) is still lacking. 
In this section, we perform qualitative coding over all \emph{130} free-form, summary-level comments from \fables\ and present a taxonomy of content selection errors (e.g., omissions) that may prove more difficult to detect than faithfulness.\footnote{Details of the annotation scheme used to analyze the comments are in \autoref{tab:annotator-comments-criteria} in the \S\ref{app:comments-section}}

\paragraph{General issues with LLM-generated summaries:} 
\autoref{tab:performance_metrics} summarizes the percentage of summaries affected by specific issues as per annotators' comments.\footnote{In two cases, \claude\ refused to merge two summaries, as they were affected by the extra information available in the front and back matter  and did not constitute a logical story. We excluded these cases from this analysis.} Our analysis shows that every LLM makes chronological errors, though these were less pronounced in models with extended context (\claude\ and \gptturbo).
All models were also criticized for omitting important information, with \claude\ being the least affected (52\%), compared to 80.8\% and  84.6\% for \gptturbo\ and \chatgpt, respectively. 
The least faithful models, \chatgpt\ and \mixtral, also both have a tendency to generate overly generic statements (38.5\%). Finally, we look also at cases where the summary was explicitly praised for being good or comprehensive. \claude\ received the most praise (48\% and 54\% respectively), while \chatgpt\ received the least (11.5\% and 15.4\% respectively).

\begin{table}[h]
\centering
\scalebox{0.95}{
\begin{tabular}{lccccc}
\toprule
 & \claude & \gptturbo & \gpt & \chatgpt & \mixtral \\
\midrule
\color{purple}{\faThumbsODown\ Chronology}  & 33.3 & 36.0 & 46.2 & 50.0 & 61.5 \\
\color{purple}{\faThumbsODown\ Omissions}   & 52.0 & 80.8 & 65.4 & 84.6 & 65.4 \\
\color{purple}{\faThumbsODown\ Factuality}\tablefootnote{Percentage of summaries where the annotator expressed specific concerns about the factuality of the entire claim set. See \S\ref{app:additional_by_book_factuality_count} for the percentage of affected claims per summary. In short, most summaries contained factual inaccuracies with only five summaries receiving 100\% of \Faithful\ labels (indicating complete factual accuracy).}   & 58.3 & 69.2 & 80.8 & 69.2 & 84.6 \\
\color{purple}{\faThumbsODown\ Overemphasis} & 20.8 & 34.6 & 19.2 & 30.8 & 46.2 \\
\color{purple}{\faThumbsODown\ Underemphasis}  & 12.5 & 23.1 & 19.2 & 38.5 & 34.6 \\
%Hallucinations    & 4.00 & 15.38 & 7.69 & 3.85 & 7.69 \\
\color{purple}{\faThumbsODown\ Vague/Generic}  & 0.0 & 23.1 & 3.9 & 38.5 & 38.5 \\
\color{purple}{\faThumbsODown\ Repetitive} & 0.0 & 11.5 & 0.0 & 7.7 & 3.9 \\
\color{purple}{\faThumbsODown\ Data-Influenced} & 0.0 & 23.1 & 19.2 & 19.2 & 34.6 \\
\midrule
\color{teal}{\faThumbsOUp\ Comprehensive}   & 54.2 & 30.8 & 38.5 & 15.4 & 34.6 \\
\color{teal}{\faThumbsOUp\ Well-done}   & 50.0 & 23.1 & 26.9 & 11.5 & 15.4 \\
\bottomrule
\end{tabular}
}
\caption{Percentage of summaries per model identified with specific issues, \textbf{based on annotator general comments} (not the claim-wise faithfulness ratings). The upper row, colored in {\color{purple}\textbf{purple}}, outlines categories of critique, whereas the lower row, in {\color{teal}\textbf{green}}, indicates categories where the models received compliments.}
\label{tab:performance_metrics}
\end{table}

\begin{figure}[h]
\centering
\includegraphics[width=1\linewidth]{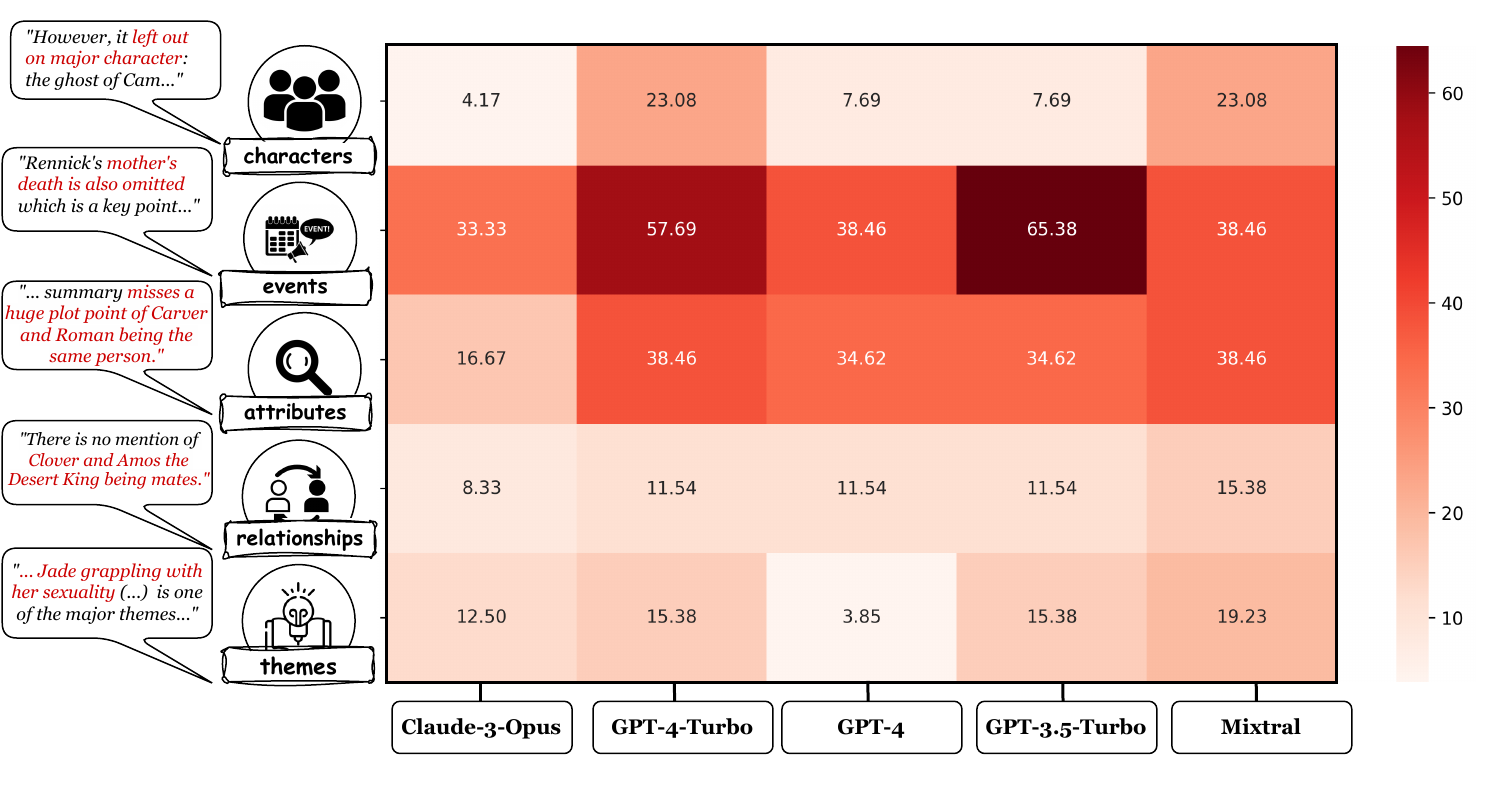}
\caption{Percentage of summaries flagged by the annotators for one of five omission errors, \textit{characters}, \textit{events}, \textit{attributes}, \textit{relationships}, and \textit{themes}, by model.}
\label{fig:omissions_errors_type_heatmap}
\end{figure}

\paragraph{Exploring omission errors:} As mentioned above, omission of key information plagues all LLM summarizers. To better understand the nature of the omission errors identified by our annotators, we categorize them into the following categories: \textit{characters}, \textit{events}, \textit{details}, \textit{relationships}, \textit{themes}.\footnote{Since annotators did not identify every specific omission, we focused on a binary classification: whether a summary was impacted by a given omission type, rather than counting the total number of omissions by type. See \autoref{tab:omissions-types-criteria} in the \S\ref{app:comments-section} for more details.} Figure~\ref{fig:omissions_errors_type_heatmap} shows a heatmap of omission errors broken down by model. A large proportion of summaries (33.3\% to 65.4\%) lack mentions of key events, creating gaps in the overall narrative, and we also note omissions of significant details about the characters, events, or objects (16.7\% to 38.5\%). Furthermore, \gptturbo\ and \mixtral\ have a tendency to entirely omit mentions of crucial characters (23.1\%).

\paragraph{Long-context models overemphasize book endings:} One interesting observation is that \claude\ and \gptturbo, which both have chunk sizes $\geq$ 100K, tend to place more emphasis on the endings of the books to the detriment of the beginning. Since these models were often provided with the entire book context during prompting, this suggests a potential issue in processing long inputs 
 \citep{Needle_In_A_Haystack, levy2024task}. This phenomenon is especially prominent with \claude, where at least 20\% of the generated summaries exhibit an overemphasis on the book's ending, compared to  7.7\% for \gptturbo\ (see examples in \autoref{tab:focus-on-the-end} in the \S\ref{app:comments-section}). We also note that the back matter of many books (e.g., author's biography, dedications, etc.) often unduly impacts all LLMs during the summarization process. We observe conflation between characters in the narrative and names in the back matter, as well as entirely hallucinated narratives; \claude\ is the only model seemingly unaffected by this additional information; see \S\ref{app:comments-section} for more analysis on this phenomenon.

\section{Related work}
\paragraph{Narrative summarization:} 
Our paper builds on prior work in narrative summarization, including short stories~\citep{wang-etal-2022-squality,subbiah2024reading}, poetry~\citep{mahbub-etal-2023-unveiling}, screenplays~\citep{chen2022summscreen}, among others.
\citet{wu2021recursively} demonstrated how an LLM can overcome long context to summarize books, like those in the BookSum~\citep{kryscinski-etal-2022-booksum} dataset.
Closely related to our work is \citet{Chang2023BooookScoreAS}, but while they focus on evaluating summary coherence (which requires only judging the model generation), we address faithfulness and content selection (which requires relating model generations back to the long source inputs).

\paragraph{Faithfulness and content selection in summarization:}
Our paper builds on prior work in evaluating hallucination and inconsistency in summarization~\citep{maynez-etal-2020-faithfulness, kryscinski-etal-2020-evaluating, Ladhak2024} which are even challenging for humans \citep{13056}.
\citet{pagnoni-etal-2021-understanding} introduce the FRANK dataset, where they use human annotations of generated summaries to produce a taxonomy of factual errors based on linguistic analysis, resembling the work of \citet{goyal-durrett-2020-evaluating} and \citet{goyal2021annotating}. 
Closest to our work, \citet{krishna-etal-2023-longeval} perform human evaluation of faithfulness on summaries of short stories, whereas we study book-length inputs.
Our exploration of omission errors is rooted in prior research on content selection~\citep{nenkova-passonneau-2004-evaluating, gillick-liu-2010-non, ladhak-etal-2020-exploring}.

\paragraph{Claim verification for evaluating summaries:}
Our paper also relates to prior work on claim verification, where claims are verified given reference to some knowledge source \citep{thorne-etal-2018-fever, wadden-etal-2020-fact, schuster-etal-2021-get}.
\citet{DBLP:conf/emnlp/MinKLLYKIZH23} propose FActScore, an LLM-based metric of factual precision in biography generation, which was expanded upon in SAFE \citep{wei2024longform}. \cite{manakul2023selfcheckgpt} propose SelfCheckGPT, which uses LLMs to evaluate the faithfulness of \textsc{GPT-3} generated texts on a dataset of Wikipedia-style passages about people.

\section{Conclusion}

We present \fables, the first large-scale human evaluation of faithfulness and content selection in book-length summarization. By recruiting annotators who had read recently-published books for enjoyment, we collect 3,158 claim-level faithfulness annotations from LLM-generated summaries of 26 narratives. This allows us to rank LLM summarizers based on faithfulness, revealing that \claude\ is the most faithful book-length summarizer, followed by \gptturbo. Next, we experiment with using LLMs for automatic claim verification. Our results expose the limitations of both retrieval and long-context understanding: LLM auto-raters cannot reliably detect \emph{unfaithful} claims, even when prompted with the full book text. Our analysis shows that unfaithful claims primarily pertain to states and events, often necessitating reasoning over extended contexts, which makes them complicated to detect for both humans and machines. Finally, we move beyond faithfulness to explore and characterize common content selection errors such as omissions of key events, attributes, and characters, as well as the over-emphasis of content from the end of the book. 

Our work on \fables\ suggests several promising directions for future work. With better auto-raters of faithfulness, we can perform fine-tuning or preference tuning on long-context language models by using the auto-raters as a scorer \citep{Tian2023FinetuningLM}, which could improve their summarization capabilities by reducing hallucination~\citep{Cao2021HallucinatedBF}. Additionally, \fables\ can be used as a dataset and protocol to meaningfully benchmark future work on novel long-context language model architectures and training objectives.

\section*{Ethical considerations}

All annotators consented to the use and publication of their annotations. The dataset excludes copyrighted texts, containing only annotations done on model-generated summary claims. Additionally, we ensured annotators received fair compensation for their contributions.
\section*{Acknowledgments}
We extend special gratitude to the Upwork annotators for their hard work, and to members from the UMass NLP lab for their feedback. This project was partially supported by awards IIS-2202506 and IIS-2312949 from the National Science Foundation (NSF) as well as an award from Adobe.

\bibliography{colm2024_conference}
\bibliographystyle{colm2024_conference}

\appendix
\section{Dataset}
\label{app:dataset-details-section}

In this section, we include further details about \fables. We list all the books used for summarization in \autoref{tab:books_in_corpus}, along with details about the authors, genre, length, publication date, and variety of English. We also detail the data preprocessing process in \S\ref{app:data-processing-subsection}.

\subsection{Data Preprocessing}
\label{app:data-processing-subsection}

\paragraph{Preprocessing books:} In order to obtain the summaries via hierarchical merging, we first purchased books from \texttt{amazon.com} in \texttt{epub} format and converted them into text files, retaining all information intact (i.e., \textit{without} removing front and back matter). We then used the Huggingface GPT-2 tokenizer\footnote{\url{https://huggingface.co/docs/transformers/en/model_doc/gpt2}} to divide the books into chunks fitting the models` context window. During our chunking step, we checked for punctuation marks to ensure that all chunks end with a complete sentence. This approach sometimes resulted in chunks being shorter than the specified size, leading to the final chunks of some books consisting only of brief snippets with meta information, which could influence the summaries. Ideally, a robust model would distinguish between supplementary information and the main storyline to produce a coherent summary. However, we observed that some models were influenced by this extra information, leading them to fabricate aspects of the story.

\paragraph{Generating summaries:} To summarize book-length documents, we adopt the hierarchical merging strategy which \citet{Chang2023BooookScoreAS} found to outperform competing approaches in terms of summary coherence. 
We employ zero-shot prompting to summarize each chunk independently. Next, we form pairs of adjacent chunk-level summaries and again use zero-shot prompting to merge each pair, incorporating added context from previously-generated merged summaries to ensure coherence and continuity (see \autoref{fig:pipeline}a). 
We generate five summaries for each book in this fashion using \chatgpt, \gpt, \gptturbo\ \citep{DBLP:journals/corr/abs-2303-08774}, \mixtral\ \citep{jiang2024mixtral}, and \claude\ \citep{Anthropic2023Claude}. All summaries were generated in February 2024 using the following checkpoints: \texttt{gpt-3.5-turbo}, \texttt{gpt-4}, \texttt{gpt-4-turbo-preview}, \texttt{Mixtral-8x7B-Instruct-v0.1}, and \texttt{claude-3-opus-20240229}. We use publicly-released code, prompts, and hyperparameters from~\citet{Chang2023BooookScoreAS} for summary generation. We further prompt \gpt\ model to extract decontextualized claims from the summaries. Examples of summaries along with extracted claims can be found in \autoref{tab:summaries-examples-claude}, \autoref{tab:summaries-examples-gptturbo}, \autoref{tab:summaries-examples-gpt}, \autoref{tab:summaries-examples-chatgpt}, and \autoref{tab:summaries-examples-mixtral}.

% BOOK DETAILS
\begin{table}[ht]
\centering
\scalebox{0.75}{ 
\begin{tabular}{>{\raggedright\arraybackslash}p{3.5cm} 
                >{\raggedright\arraybackslash}p{2.8cm} 
                >{\centering\arraybackslash}p{1.2cm} 
                >{\raggedright\arraybackslash}p{3cm} 
                >{\raggedright\arraybackslash}p{1.5cm} 
                >{\raggedright\arraybackslash}p{2cm} 
                >{\raggedright\arraybackslash}p{2cm}}
\toprule
\textsc{Title} & \textsc{Author} & \textsc{Gender} & \textsc{Genre} & \textsc{Length} & \textsc{Publication} & \textsc{Lang} \\
\midrule
\textit{A Haunting on the Hill} & Elizabeth Hand 
& \textsc{F} 
& horror, Gothic & 117,577 & Oct 3, 2023 & AmE \\
\textit{Agency for Scandal} & Laura Wood & \textsc{F} & historical fiction, mystery, romance & 116,809 & Jan 5, 2023 & BrE \\
\textit{Divine Rivals} & Rebecca Ross  & \textsc{F} & fantasy, romance, young adult & 137,616 & Apr 4, 2023 & AmE \\
\textit{Fairytale of New York} & Zoë Folbigg & \textsc{F} & romance & 134,369 & Aug 28, 2023 & BrE \\
\textit{Flawless} & Elsie Silver & \textsc{F} & romance & 119,580 & Jun 24, 2022 & CanE \\
\textit{Fourth Wing} & Rebecca Yarros & \textsc{F} & fantasy & 243,965 & May 2, 2023 & AmE \\
\textit{Modern Divination} & Isabel Agajanian & \textsc{F} & urban fantasy & 167,568 & Jan 30, 2023 & AmE \\
\textit{Only for the Week} & Natasha Bisho & \textsc{F} & African American romance & 87,056 & May 11, 2023 & AmE \\
\textit{Pet} & Catherine Chidgey & \textsc{F} & thriller, mystery & 124,679 & July 13, 2023 & NZE \\
\textit{Romantic Comedy} & Curtis Sittenfeld & \textsc{F} & romance & 116,560 & Apr 4, 2023 & AmE \\
\textit{Same Time Next Year} & Tessa Bailey & \textsc{F} & romance & 49,762 & Dec 1, 2023 & AmE \\
\textit{She is a Haunting} & Trang Thanh Tran & \textsc{F} & romance & 106,659 & Feb 28, 2023 & AmE \\
\textit{Six Scorched Roses} & Carissa Broadbent & \textsc{F} & fantasy romance & 54,481 & Mar 21, 2023 & AmE \\
\textit{Sorrow and Bliss} & Meg Mason & \textsc{F} & mental health & 112,468 & Sep 2, 2020 & AusE \\
\textit{The Atonement Murders} & Jenifer Ruff & \textsc{F} & mystery, thriller & 105,493 & Apr 14, 2023 & AmE \\
\textit{The Guest} & Emma Cline & \textsc{F} & thriller & 89,977 & May 16, 2023 & AmE \\
\textit{The Marriage Act} & John Marrs & \textsc{M} & thriller, mystery, dystopia & 135,901 & Jan 19, 2023 & BrE \\
\textit{The Spy Coast} & Tess Gerritsen & \textsc{F} & mystery, thriller & 128,918 & Nov 1, 2023 & AmE \\
\textit{The Wager} & David Grann & \textsc{M} & nonfiction, historical & 156,022 & Apr 18, 2023 & AmE \\
\textit{The White Lady} & Jacqueline Winspear & \textsc{F} & historical fiction, mystery & 126,051 & Mar 21, 2023 & BrE \\
\textit{This Impossible Brightness} & Jessica Bryant Klagmann & \textsc{F} & mystery, fantasy & 119,115 & Feb 1, 2024 & AmE \\
\textit{Viciously Yours} & Jamie Applegate Hunter & \textsc{F} & fantasy, romance & 86,774 & Jan 23, 2024 & AmE \\
\textit{Weyward} & Emilia Hart & \textsc{F} & historical fiction, fantasy & 128,728 & Feb 2, 2023 & AusE \\
\textit{Wildfire} & Hannah Grace & \textsc{F} & romance, sport & 140,060 & Oct 3, 2023 & BrE \\
\textit{Yellowface} & R.F. Kuang & \textsc{F} & thriller & 114,346 & May 25, 2023 & AmE\\
\textit{You, Again} & Kate Goldbeck & \textsc{F} & romance & 137,624 & Sep 12, 2023 & AmE \\
\bottomrule
\end{tabular}
} % End scalebox
\caption{Details of the 26 books used for summaries. Length of each book is provided in tokens as computed with \texttt{tiktoken}.}
\label{tab:books_in_corpus}
\end{table}

% claude
\begin{table}[ht]
\centering
\footnotesize
\scalebox{0.95}{
\begin{tabular}{@{}p{0.05\linewidth}p{0.95\linewidth}@{}}
\toprule
 & \claude \\ 
\midrule
\addlinespace
\multirow{15}{*}{\rotatebox{90}{Summary}} & \textit{Iris Winnow, an aspiring writer, works at the Oath Gazette newspaper in the city of Oath, competing with her rival Roman Kitt for a columnist position. Iris's brother Forest is fighting in a war between the gods Enva and Dacre. She hasn't heard from him in months and decides to become a war correspondent to find him. Iris and Roman develop feelings for each other through letters they exchange via enchanted typewriters, not realizing they are writing to each other. Roman, engaged to Elinor Little through an arranged marriage by his father, breaks off the engagement and follows Iris to the war front. In the town of Avalon Bluff near the front lines, Iris stays with Marisol, who runs a bed and breakfast, and befriends fellow correspondent Attie. Roman arrives and he and Iris fall in love, getting married just before an attack on the town by Dacre's forces. During the attack, Iris is saved by a disguised soldier who she later realizes is her brother Forest. Forest, who was supposed to have died from his wounds but was healed by Dacre, deserted and has been fighting for the enemy. In the chaos, Iris is separated from Roman, who is badly wounded. Iris and Forest return to Oath, where Iris learns the full story of Forest's service to Dacre. She discovers a letter from Roman in her room, confirming he knew her identity while writing to her. Meanwhile, the god Dacre finds the dying Roman on the battlefield and, sensing his strong spirit, decides to take him to heal his wounds and have him fight for his army. The story ends with Dacre turning his sights towards finding Enva in the east.} \\
\addlinespace
\multirow{25}{*}{\rotatebox{90}{Extracted Claims}} & 1. Iris Winnow works at the Oath Gazette newspaper in the city of Oath. \newline
2. Iris Winnow is competing with Roman Kitt for a columnist position at the Oath Gazette. \newline
3. Iris Winnow's brother, Forest, is involved in a war between the gods Enva and Dacre. \newline
4. Iris Winnow decides to become a war correspondent after not hearing from her brother, Forest, for months. \newline
5. Iris Winnow and Roman Kitt develop feelings for each other through letters they exchange via enchanted typewriters. \newline
6. Iris Winnow and Roman Kitt do not initially realize they are writing letters to each other. \newline
7. Roman Kitt was engaged to Elinor Little through an arranged marriage by his father. \newline
8. Roman Kitt breaks off his engagement to Elinor Little and follows Iris Winnow to the war front. \newline
9. Iris Winnow stays with Marisol at a bed and breakfast in the town of Avalon Bluff near the front lines. \newline
10. Iris Winnow befriends fellow correspondent Attie in Avalon Bluff. \newline
11. Roman Kitt arrives at Avalon Bluff and he and Iris Winnow fall in love. \newline
12. Roman Kitt and Iris Winnow get married just before an attack on Avalon Bluff by Dacre's forces. \newline
13. During the attack on Avalon Bluff, Iris Winnow is saved by a soldier who she later realizes is her brother, Forest. \newline
14. Forest was supposed to have died from his wounds but was healed by Dacre. \newline
15. Forest deserted from his original side and has been fighting for Dacre. \newline
16. Iris Winnow is separated from Roman Kitt, who is badly wounded during the attack on Avalon Bluff. \newline
17. Iris Winnow and Forest return to Oath after the attack. \newline
18. Iris Winnow learns the full story of Forest's service to Dacre once they return to Oath. \newline
19. Iris Winnow finds a letter from Roman Kitt in her room, which confirms he knew her identity while writing to her. \newline
20. Dacre finds the dying Roman Kitt on the battlefield and decides to take him to heal his wounds. \newline
21. Dacre plans for the healed Roman Kitt to fight for his army. \newline
22. The story ends with Dacre planning to find Enva in the east. \\ 
\addlinespace
\bottomrule
\addlinespace
\end{tabular}
}
\caption{Example of a summary produced by \claude\ along with the extracted set of claims for ``Divine Rivals,'' a novel by Rebecca Ross. Examples by the other models can be found in \autoref{tab:summaries-examples-gptturbo}, \autoref{tab:summaries-examples-gpt}, \autoref{tab:summaries-examples-chatgpt} and \autoref{tab:summaries-examples-mixtral}.}
\label{tab:summaries-examples-claude}
\end{table}

% gpt-4-turbo
\begin{table}[ht]
\centering
\footnotesize
\scalebox{0.95}{
\begin{tabular}{@{}p{0.05\linewidth}p{0.95\linewidth}@{}}
\toprule
 & \gptturbo \\ 
\midrule
\addlinespace
\multirow{15}{*}{\rotatebox{90}{Summary}} & \textit{In the epic narrative of "Divine Rivals," we are introduced to Iris Winnow, a tenacious young woman propelled into the role of a war correspondent amidst a devastating conflict influenced by the ancient deities Enva and Dacre. Iris's journey into the heart of battle is not just professional but deeply personal, fueled by the disappearance of her beloved brother, Forest, who had enlisted to fight, and the anguish over her mother's demise. Her quest for truth and understanding is set against the backdrop of the war-torn Avalon Bluff, close to the war's front lines, where she encounters the stark realities of conflict, from the silent hope of a garden amidst destruction to the horror of wounded soldiers and the terror of artillery fire. Roman Kitt, previously Iris's rival at the Oath Gazette, emerges as a significant figure in her narrative, joining her as a fellow correspondent. Despite their history of competitive antagonism, their relationship evolves, highlighted by Roman's mysterious decision to leave behind a promising career and a broken engagement to be by Iris's side. Their dynamic is further complicated and enriched by their shared experiences on the battlefield and Roman's unspoken feelings for Iris, culminating in a poignant moment of sacrifice during a trench attack where Roman shields Iris with his body from a grenade explosion. Adding a magical element to the story, Iris maintains a correspondence with Carver, a mysterious figure, through enchanted typewriters. These letters offer Iris solace and a metaphysical connection that transcends the chaos surrounding her, underscoring themes of hope and connection. As the narrative unfolds, shocking revelations come to light. Forest, Iris's brother, believed to have been fighting for Enva, is discovered to have been forcefully conscripted by Dacre after a near-death experience. This discovery unfolds through a dramatic reunion where Forest saves Iris from one of Dacre's deadly gas attacks, only to reveal his changed allegiance by deceiving her and leaving Roman behind, presumed dead or captured. The heart of the story explores Iris and Roman's profound relationship, against the backdrop of war and divine manipulation. Moments from their past, including their secret wedding in a garden before a devastating attack by Dacre's forces, and Roman's heartfelt letter confessing his love, serve as poignant reminders of what they fight for. This emotional core is juxtaposed with the chaos of the present, wherein Iris must confront the betrayal of her brother and the uncertainty of Roman's fate. "Divine Rivals" delves into themes of love, loyalty, and the harrowing choices faced by individuals caught in the machinations of gods and war. Iris, within the turmoil of her personal and professional world, remains determined to uncover the truth, seeking to reclaim agency and hope amidst despair. Her journey is not just a search for her brother and Roman but an exploration of the resilience of the human spirit in the face of divine conflicts and the devastation of war. Through Iris's eyes, the story weaves a complex tapestry of human emotions, the indomitable will to survive, and the quest for truth in a reality marred by capricious gods and endless conflict.} \\
\addlinespace
\multirow{25}{*}{\rotatebox{90}{Extracted Claims}} & 
1. Iris Winnow is the main character in the book "Divine Rivals." \newline
2. Iris Winnow becomes a war correspondent during a conflict influenced by the deities Enva and Dacre. \newline
3. Iris's brother, Forest, has disappeared after enlisting to fight in the war.
4. Iris is devastated by her mother's death. \newline
5. Iris's journey takes her to Avalon Bluff, a location close to the war's front lines. \newline
6. Iris encounters the realities of war, including wounded soldiers and artillery fire. \newline
7. Roman Kitt is a significant character in the narrative and joins Iris as a fellow war correspondent. \newline
8. Roman Kitt was Iris's rival at the Oath Gazette before they became colleagues. \newline
9. Their relationship evolves, despite a history of competitive antagonism. \newline
10. Roman Kitt leaves a promising career and a broken engagement to join Iris. \newline
11. Roman Kitt shields Iris from a grenade explosion during a trench attack. \newline
12. Iris maintains a magical correspondence with a mysterious figure named Carver through enchanted typewriters. \newline
13. The letters from Carver offer Iris solace and a metaphysical connection. \newline
14. Forest was believed to be fighting for deity Enva, but is revealed to have been forcefully conscripted by deity Dacre. \newline
15. This revelation comes after Forest saves Iris from a gas attack by Dacre and then deceives her by leaving Roman behind. \newline
16. Iris and Roman had a secret wedding in a garden before an attack by Dacre's forces. \newline
17. Roman writes a letter confessing his love for Iris. \newline
18. Iris must confront her brother's betrayal and uncertainty over Roman's fate. \newline
19. Iris is determined to uncover the truth and reclaim agency and hope amidst despair. \newline
20. Iris's journey is a search for her brother and Roman, and an exploration of human resilience in the face of conflict and divine manipulation. \newline
21. The story of "Divine Rivals" explores themes of love, loyalty, and the difficult choices faced by individuals in times of war. \newline \\ 
\addlinespace
\bottomrule
\addlinespace
\end{tabular}
}
\caption{Example of a summary produced by \gptturbo\ along with the extracted set of claims for ``Divine Rivals,'' a novel by Rebecca Ross. Examples by the other models can be found in \autoref{tab:summaries-examples-claude}, \autoref{tab:summaries-examples-gpt}, \autoref{tab:summaries-examples-chatgpt} and \autoref{tab:summaries-examples-mixtral}.}
\label{tab:summaries-examples-gptturbo}
\end{table}

%gpt-4
\begin{table}[ht]
\centering
\footnotesize
\scalebox{0.85}{
\begin{tabular}{@{}p{0.05\linewidth}p{0.95\linewidth}@{}}
\toprule
 & \gpt \\ 
\midrule
\addlinespace
\multirow{15}{*}{\rotatebox{90}{Summary}}& \textit{Divine Rivals is an intricate narrative set in a dystopian world at war, invoked by two resurrected rival deities, Enva and Dacre. The plot centres around Iris Winnow, a sensitive and tenacious employee at the Oath Gazette newspaper in the city of Oath, and her professional rival Roman Kitt. Iris's uneventful life spirals into chaos with her brother, Forest, joining the gods' war, leaving her with the responsibility of their alcoholic mother, Aster. Caught in a whirlpool of personal and professional rivalry, Iris's bond with Roman deepens in unexpected ways when Roman anonymously receives and responds to Iris's emotionally raw letters intended for Forest. Iris uses a magical vintage typewriter to send these letters, leading to an anonymous correspondence that amplifies their unspoken connection. Roman's social standing is a stark contrast to Iris's modest background; he is affianced to a high society woman, Elinor Little, and faces the challenges of striking a balance between his family's aspirations and his new role at Oath Gazette.  Following their mother's sudden passing, Iris impulsively resigns from her job at Oath Gazette. Determined to locate Forest and report on the escalating gods' war, she accepts a position as a war correspondent with the Inkridden Tribune newspaper. Iris embarks on a perilous journey to Avalon Bluff, a town near the war front, accompanied by another correspondent, Thea "Attie" Attwood. Their hostess in the eerie town, Marisol Torres, warns the pair about the regular threats from different mythical creatures such as hounds and eithrals that desolate their surroundings. Roman's arrival in Avalon Bluff reignites their convoluted relationship. He assists Iris in documenting the harrowing realities of the war while managing the growing tension between them. As Roman becomes severely injured during a horrific encounter at the warfront, Iris, who has grown deeply involved in reporting the soldiers' personal narratives, fights her way to ensure his safety. This life-threatening event leads Iris to recognise Roman as Carver, her mysterious pen pal, triggering an intense turmoil of guilt and suspicion over Roman's subterfuge. Amid her growing concern for Roman's fate and her conflicted feelings about his deception, a shocking twist reveals that Forest, previously presumed dead, has survived and fought for Dacre, the enemy deity. Further ensuing chaos sees Roman disappear, leaving Iris trapped between her obligations towards her effectively traitorous brother and the desperate search for her missing love interest.  "Divine Rivals" interweaves Iris's personal story of resilience and growth against the backdrop of an escalating war. As she metamorphoses from a correspondent to a key player in the war, Iris embodies the human spirit's ability to endure and adapt amidst adversity. The narrative encapsulates the strength of character, familial obligations, professional rivalries, societal norms, and profound transformations at the heart of the story. Despite the evolving relationships and hardships, Iris pledges to remain by Forest's side and continue her search for Roman. The story of "Divine Rivals" is a testament to human endurance, love, and loyalty amidst a surreal world at war.} \\ 
\addlinespace
\multirow{25}{*}{\rotatebox{90}{Extracted Claims}}& 
1. Divine Rivals is a narrative set in a dystopian world at war. \newline
2. The war in the world of "Divine Rivals" is invoked by two resurrected rival deities, Enva and Dacre. \newline
3. The story's plot focuses on Iris Winnow, an employee at the Oath Gazette newspaper in the city of Oath. \newline
4. Iris Winnow has a professional rival named Roman Kitt. \newline
5. Iris Winnow's brother, Forest, joins the god's war. \newline
6. Iris Winnow is left with the responsibility of their alcoholic mother, Aster, after 7. Forest joins the war. \newline
8. Iris Winnow's bond with Roman Kitt deepens when Roman anonymously receives and responds to Iris's letters intended for Forest. \newline
9. Iris Winnow uses a magical vintage typewriter to send letters to Roman Kitt. \newline
10. Roman Kitt is engaged to a high society woman named Elinor Little. \newline
11. Roman Kitt faces challenges balancing his family's aspirations and his role at the Oath Gazette. \newline
12. Iris Winnow resigns from her job at the Oath Gazette after their mother's sudden death. \newline
13. Iris Winnow accepts a position as a war correspondent with the Inkridden Tribune newspaper. \newline
14. Iris Winnow journeys to Avalon Bluff, a town near the war front, with another correspondent named Thea "Attie" Attwood. \newline
15. In Avalon Bluff, Marisol Torres warns Iris Winnow and Thea Attwood about the regular threats from mythical creatures. \newline
16. Roman Kitt arrives in Avalon Bluff and assists Iris Winnow in documenting the war. \newline
17. Roman Kitt is severely injured during an encounter at the warfront. \newline
18. Iris Winnow realizes Roman Kitt is Carver, her mysterious pen pal, after he is severely injured. \newline
19. Iris Winnow feels guilt and suspicion over Roman Kitt's deception. \newline
20. It is revealed that Forest, previously assumed dead, survived and fought for Dacre, the enemy deity. \newline
21. Roman Kitt disappears, leaving Iris Winnow torn between her obligations towards her brother and the search for Roman. \newline
22. Iris Winnow evolves from a correspondent to a key player in the war. \newline
23. Iris Winnow pledges to stay by her brother's side and continue her search for Roman Kitt. \newline
24. The story of "Divine Rivals" is a testament to human endurance, love, and loyalty during a war. \\ 
\addlinespace
\bottomrule
\addlinespace
\end{tabular}
}
\caption{Example of a summary produced by \gpt\ along with the extracted set of claims for ``Divine Rivals,'' a novel by Rebecca Ross. Examples by the other models can be found in \autoref{tab:summaries-examples-claude}, \autoref{tab:summaries-examples-gptturbo}, \autoref{tab:summaries-examples-chatgpt} and \autoref{tab:summaries-examples-mixtral}.}
\label{tab:summaries-examples-gpt}
\end{table}

%chatgpt
\begin{table}[ht]
\centering
\footnotesize
\scalebox{0.95}{
\begin{tabular}{@{}p{0.05\linewidth}p{0.95\linewidth}@{}}
\toprule
 & \chatgpt \\ 
\midrule
\addlinespace
\multirow{15}{*}{\rotatebox{90}{Summary}} & \textit{In the city of Oath, siblings Iris and Forest Winnow face separation as Forest departs for war, leaving Iris to care for their mother. As Iris grapples with loneliness and worry, a fierce rivalry emerges at the Oath Gazette between Iris and Roman Kitt, two ambitious journalists. The backdrop of an impending war between the goddess Enva and god Dacre adds tension and intrigue to their lives. Iris, driven by a desire to prove herself in a male-dominated field, finds herself entangled in personal challenges while navigating her complicated relationship with Roman. As the war escalates and mysterious creatures surface, Iris and Roman become immersed in a world of gods, magic, and journalism, facing both personal and professional dilemmas. Iris decides to resign from her job to embark on a journey to find her brother Forest, who becomes entwined in the conflict. Roman, haunted by his past and familial obligations, struggles to reconcile his feelings for Iris while maintaining his journalistic integrity. Their paths eventually converge in Avalon Bluff, where Iris becomes a war correspondent equipped with a mysterious typewriter with a significant history. As Iris and Roman navigate the chaos of war, their interactions reveal deeper emotions and motivations, leading to a fateful meeting at a high-end restaurant. Attie, a friend of Iris, discloses her talent for playing the violin amidst the turmoil of war, adding a layer of depth to the narrative. Roman's vulnerability and inner turmoil are brought to the forefront when he is saved by Iris during an air raid, further strengthening their bond. As they confront personal and professional challenges amidst the impending battle, Iris reunites with Forest, only to discover his involvement with the opposing forces. Forest, torn between protecting Iris and confronting his past actions, adds a layer of complexity and tension to the story. Amidst the impending attack by Dacre, Iris and Roman make a decision to stay and assist in Avalon Bluff, preparing for the inevitable confrontation. The intertwining destinies of the characters unravel themes of sacrifice, resilience, and the search for identity in a world plagued by war and loss. As truths unfold, loyalty is tested, blurring the lines between good and evil. Characters like Dacre and Enva reveal sinister plans, setting the stage for a complex and emotional journey fraught with uncertainty and conflict. Through grief, rivalry, and love, Iris and Roman confront their vulnerabilities and ambitions, ultimately reshaping their destinies amidst the backdrop of war and its significant consequences.} \\ 
\addlinespace
\multirow{25}{*}{\rotatebox{90}{Extracted Claims}} & 
1. Iris and Forest Winnow are siblings in the city of Oath.  \newline
2. Forest Winnow departs for war and Iris stays behind to care for their mother. \newline
3. Iris struggles with feelings of loneliness and worry. \newline
4. At the Oath Gazette, a fierce rivalry develops between Iris and Roman Kitt, both of whom are ambitious journalists. \newline
5. An impending war between the goddess Enva and god Dacre adds tension and intrigue to Iris and Roman's lives. \newline
6. Iris aims to prove herself in the male-dominated field of journalism. \newline
7. Iris faces personal challenges while navigating her complicated relationship with Roman. \newline
8. As the war intensifies, Iris and Roman encounter mysterious creatures. \newline
9. Iris resigns from her job at the Oath Gazette to search for her brother Forest, who is involved in the war. \newline
10. Roman Kitt has a troubled past and family obligations that make it difficult for him to balance his feelings for Iris and his journalistic integrity. \newline
11. Iris and Roman's paths intersect in Avalon Bluff, where Iris becomes a war correspondent. \newline
12. Iris uses a typewriter with a significant past while working as a war correspondent. \newline
13. Iris and Roman's interactions during the war reveal deeper emotions and motivations. \newline
14. Iris and Roman have a significant meeting at a high-end restaurant. \newline
15. Attie, a friend of Iris, is talented at playing the violin. \newline
16. Roman's vulnerability is revealed when Iris saves him during an air raid. \newline
17. Iris reunites with Forest during the war, discovering his involvement with the opposing forces. \newline
18. Forest is torn between his desire to protect Iris and confronting his past actions. \newline
19. Amidst the approaching attack by Dacre, Iris and Roman decide to stay and help in Avalon Bluff. \newline
20. The characters' intertwining destinies reveal themes of sacrifice, resilience, and identity search in a war-torn world. \newline
21. The story explores blurred lines between good and evil as loyalties are tested. \newline
22. Characters like Dacre and Enva have sinister plans. \newline
23. Iris and Roman confront their vulnerabilities and ambitions, leading to a change in their destinies amid the backdrop of war.\\ 
\addlinespace
\bottomrule
\addlinespace
\end{tabular}
}
\caption{Example of a summary produced by \chatgpt\ along with the extracted set of claims for ``Divine Rivals,'' a novel by Rebecca Ross. Examples by the other models can be found in \autoref{tab:summaries-examples-claude}, \autoref{tab:summaries-examples-gptturbo}, \autoref{tab:summaries-examples-gpt} and \autoref{tab:summaries-examples-mixtral}.}
\label{tab:summaries-examples-chatgpt}
\end{table}

%mixtral 
\begin{table}[ht]
\centering
\footnotesize
\scalebox{0.85}{
\begin{tabular}{@{}p{0.05\linewidth}p{1.1\linewidth}@{}}
\toprule
 & \mixtral \\ 
\midrule
\addlinespace
\multirow{25}{*}{\rotatebox{90}{Summary}} & \textit{In a world torn by a devastating war between two divine families, the Skywards and the Underlings, "Divine Rivals" follows the journey of Iris Winnow, an orphaned junior reporter at the Oath Gazette in the city of Oath, as she strives to become a renowned journalist and uncover the truth about the war that took her mother and brother, Forest, from her. The Skywards and Underlings' conflict has left the realm of Cambria in chaos, where magic and mythology intertwine with the mortal world, and gods walk among humans. Iris' dream of becoming a great journalist is fueled by her love for her late brother and her desire to understand why he joined the war. She faces fierce competition from her professional rival, Roman Kitt, a talented storyteller hailing from a wealthy family. While Iris longs to write about significant issues, Roman, who shares a complicated background with Iris, excels in storytelling and is being pressured by his father to join the family business. Zeb Autry, the editor of the Oath Gazette, is reluctant to cover the war, much to Iris' dismay. Iris' family is burdened with financial difficulties and personal struggles after losing her mother and brother to the war. Driven by her determination to uncover the truth about the war and her late brother's involvement, Iris finds solace in her resolve. The Inkridden Tribune, a rival newspaper, announces an opportunity for war correspondents, enticing Iris to compete with Roman for the position. Iris's journey is further complicated by her mysterious correspondence with an unknown person who leaves her letters in her wardrobe, forming a unique bond. As Iris competes for the position, she reflects on her past and her friendship with Forest, who introduced her to the world of journalism. Their differing views on the war and the Skyward goddess, Enva, drove them apart. Iris wanted to highlight the bravery of soldiers, while Forest aimed to reveal Enva's manipulation of young minds to fuel the war. Despite the challenges, Iris remains focused on her goal and finally manages to land a job as a war correspondent for the Inkridden Tribune. She embarks on a dangerous journey to cover the war, facing various dangers and hardships along the way. Iris continues to write letters to the mysterious stranger, who becomes her confidante and support system during her perilous journey. Meanwhile, the town of Avalon Bluff prepares for an impending battle against an unknown enemy - Dacre, the god of the Underlings. Iris, Roman, and other companions gather in Avalon Bluff to aid the town's defense, as tensions rise with the theft of emergency supply kits and Keegan's distracted behavior. Amidst the chaos, Iris and Roman rekindle their connection, and Roman proposes to Iris, offering a moment of hope in the face of devastation. As the group prepares for battle, Iris witnesses firsthand the horrors of war, including the loss of innocent lives and the destruction of her childhood home, Clover Hill. Despite the pain, Iris continues to report on the war, using her words to both document the battles and search for Forest. "Divine Rivals" is a compelling narrative that explores themes of identity, courage, and the power of words in the face of war. Iris' unwavering determination to make a difference by writing about the truth, seeking answers about her brother, and making her mark as a journalist resonates throughout the story. The author extends gratitude towards the individuals who helped her publish this captivating tale, acknowledging her literary agent, film agent, subrights team, editors, cover designer, copy editor, and fellow authors.} \\
\addlinespace
\multirow{35}{*}{\rotatebox{90}{Extracted Claims}} & 
1. Divine Rivals is set in a world affected by a war between two divine families, the Skywards and the Underlings. \newline
2. The protagonist of "Divine Rivals" is Iris Winnow, an orphan who works as a junior reporter at the Oath Gazette in the city of Oath. \newline
3. Iris Winnow's aspiration to become a renowned journalist is driven by her desire to uncover the truth about the war that resulted in the death of her mother and brother, Forest. \newline
4.The conflict between the Skywards and the Underlings has caused chaos in the realm of Cambria, a place where magic and mythology interact with the mortal world. \newline
5. Iris Winnow has a professional rival named Roman Kitt, a skilled storyteller from a wealthy family. \newline
6. Roman Kitt is under pressure from his father to join the family business. \newline
7. Zeb Autry, the editor of the Oath Gazette, does not want to cover the war, against Iris Winnow's wishes. \newline
8. After the loss of her mother and brother, Iris Winnow's family is facing financial and personal struggles. \newline
9. The Inkridden Tribune, a competing newspaper, advertises a job for war correspondents, leading both Iris Winnow and Roman Kitt to compete for the position. \newline
10. Iris Winnow maintains a mysterious correspondence with an unidentified individual who leaves her letters in her wardrobe. \newline
11. Iris Winnow's brother, Forest, introduced her to journalism. \newline
12. Iris Winnow and her brother Forest had differing views on the war and the Skyward goddess, Enva. \newline
13. Despite various challenges, Iris Winnow achieves her goal of becoming a war correspondent for the Inkridden Tribune. \newline
14. Iris Winnow embarks on a dangerous journey to cover the war. \newline
15. Iris Winnow continues to write letters to the unknown person who becomes her confidante during her journey. \newline
16. The town of Avalon Bluff is preparing for a battle against an unknown enemy. \newline
17. Dacre, the god of the Underlings. \newline
18. Amidst the chaos, Iris Winnow and Roman Kitt rekindle their connection, and Roman Kitt proposes to Iris Winnow. \newline
19. Iris Winnow experiences the horrors of war firsthand, including the loss of innocent lives and the destruction of her childhood home, Clover Hill. \newline
20. Despite the hardships, Iris Winnow continues to report on the war, using her writing to document the battles and search for her brother, Forest. \newline
21. Divine Rivals explores themes of identity, courage, and the power of words during war. \newline
22. Iris Winnow's determination to make a difference by writing about the truth and seeking answers about her brother is a central theme in "Divine Rivals". \newline
23. The author of "Divine Rivals" acknowledges her literary agent, film agent, subrights team, editors, cover designer, copy editor, and fellow authors for their assistance in publishing the book.
\\
\addlinespace
\bottomrule
\addlinespace
\end{tabular}
}
\caption{Example of a summary produced by \mixtral\ along with the extracted set of claims for ``Divine Rivals,'' a novel by Rebecca Ross. Examples by the other models can be found in \autoref{tab:summaries-examples-claude}, \autoref{tab:summaries-examples-gptturbo}, \autoref{tab:summaries-examples-gpt} and \autoref{tab:summaries-examples-chatgpt}.}
\label{tab:summaries-examples-mixtral}
\end{table}

%%%%%%%%%%%%%%%%%%%%%%%%%%%%%%%%%%%%%%%%%%%%%%%%%%%%%%%%%%%%
%%%%%%%%%%%%%%%  PROMPTS   %%%%%%%%%%%%%%%%%%%%%%%%%%%%%%%%%
%%%%%%%%%%%%%%%%%%%%%%%%%%%%%%%%%%%%%%%%%%%%%%%%%%%%%%%%%%%%

\section{Prompts}
\label{app:prompt-section}
In this section, we included all prompts used for our experiments:
(1) \textit{claim extraction}, (2) \textit{automatic evaluation} in \autoref{tab:prompt_templates_claim_eval}.

% PROMPT TABLE
\begin{table}[h]
\small
    \setlength{\tabcolsep}{4pt}
    \centering
    \resizebox{0.7\columnwidth}{!}{
    \begin{tabular}{c p{9cm}}
    
    \toprule
    & \multicolumn{1}{c}{\bf Claim Extraction Template} \\
    \midrule
     \noalign{\vskip 1mm}
     & \texttt{You are trying to verify the faithfulness of statements made in a given summary of a book against the actual text of the book. To do so, you first need to break the summary into a set of "atomic claims", each of which will then be passed to a human who will read the book and verify if the claim is true or not. Each atomic claim must be fully understandable without any other context from the summary (e.g., all entities must be referred to by name, not pronoun), and they must be situated within relevant temporal, location, and causal context whenever possible. Try to keep each atomic claim to a maximum of 2 sentences. Each atomic claim is separated with '- '.} \\
     \noalign{\vskip 2mm}
     & \texttt{Summary:} \\
     \noalign{\vskip 2mm}
     & \texttt{List of atomic claims:} \\
     \noalign{\vskip 1mm}
     \bottomrule
    \toprule
         & \multicolumn{1}{c}{\bf Evaluation Template} \\
    \midrule
     \noalign{\vskip 1mm}
     & \texttt{You are provided with a context and a statement. Your task is to carefully read the context and then determine whether the statement is true or false. Use the information given in the context to make your decision.}\\
     \noalign{\vskip 2mm}
     & \texttt{Context:} \\
     \noalign{\vskip 2mm}
     & \texttt{Statement:} \\
    \noalign{\vskip 2mm}
     & \texttt{Question: Based on the context provided, is the above statement True or False?} \\
    \noalign{\vskip 2mm}
     & \texttt{Answer:} \\
     \noalign{\vskip 1mm}
    \bottomrule
    \end{tabular}
    }
    \caption{Prompt templates used for \textsc{Claim Extraction} and \textsc{Evaluation Extraction}.} 
    \label{tab:prompt_templates_claim_eval}
\end{table}

%%%%%%%%%%%%%%%%%%%%%%%%%%%%%%%%%%%%%%%%%%%%%%%%%%%%%%%%%%%%
%%%%%%%%%%%%%%%  HUM EVAL %%%%%%%%%%%%%%%%%%%%%%%%%%%%%%%%%
%%%%%%%%%%%%%%%%%%%%%%%%%%%%%%%%%%%%%%%%%%%%%%%%%%%%%%%%%%%%

%%%%%%%%% Annotators' Instructions %%%%%%%%%%%%%%%%%%%%%%%%%
\section{Human Annotations}
\label{app:hum-eval-section}

\begin{figure}[htbp]
\centering
\includegraphics[width=1\linewidth]{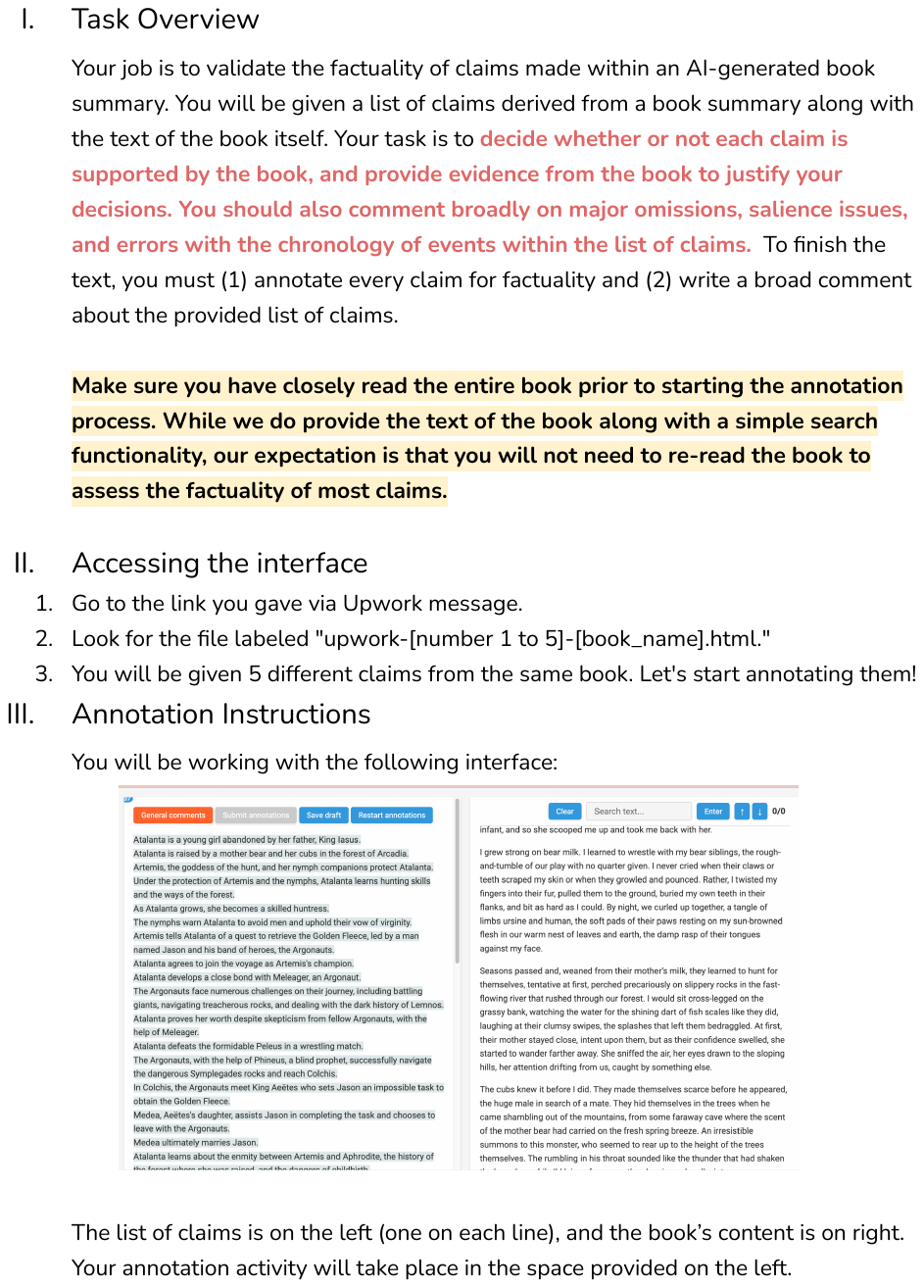}
\caption{Instructions for annotation task described in \S\ref{sec:data_method}}.
\label{fig:annot_instruction}
\end{figure}

\begin{figure}[htbp]
\centering
\includegraphics[width=1\linewidth]{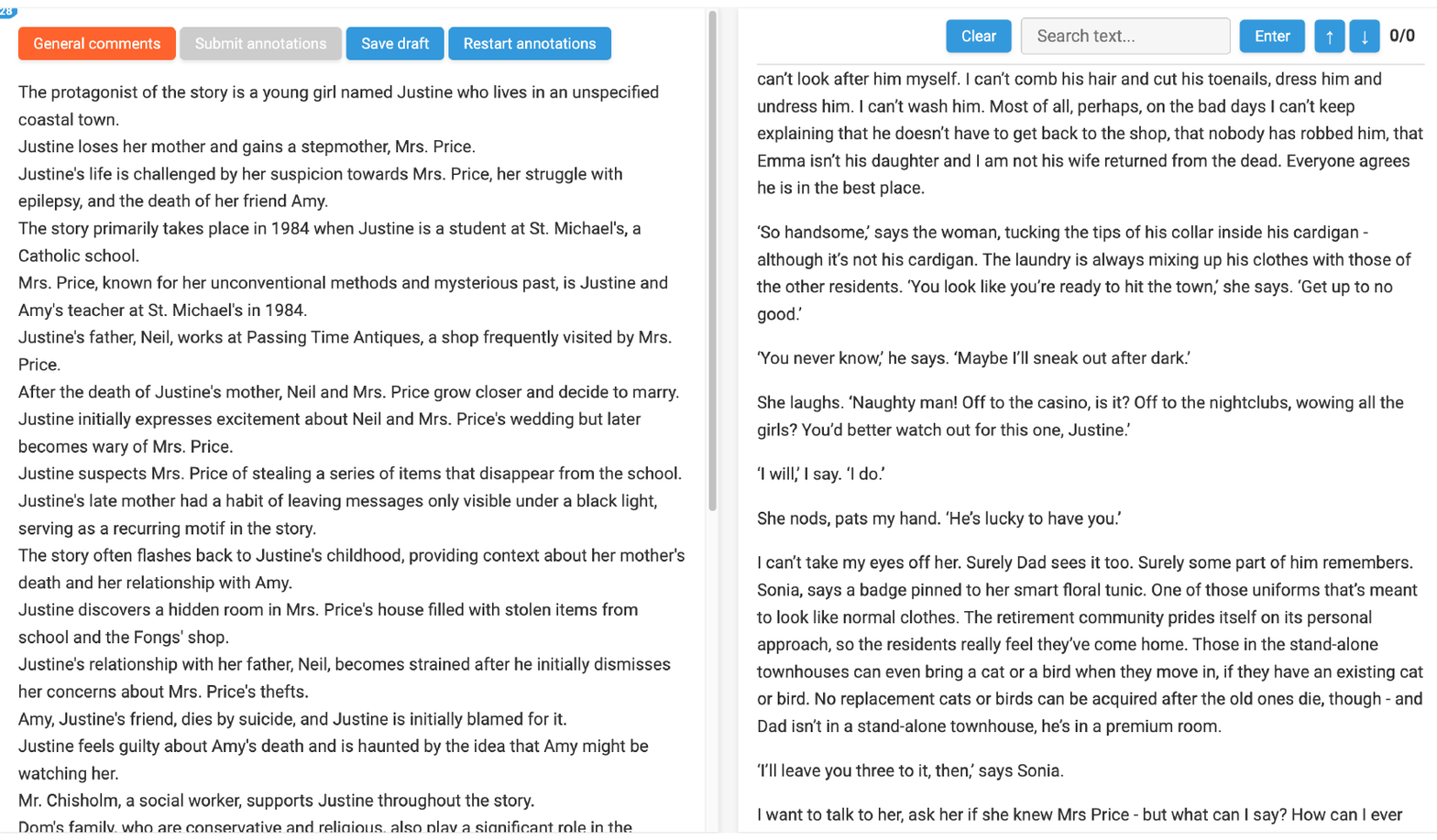}
\caption{Screenshot of the interface for the annotation task described in \S\ref{sec:data_method}.}
\label{fig:annot_interface}
\end{figure}

\begin{figure}[htbp]
\centering
\includegraphics[width=0.8\linewidth]{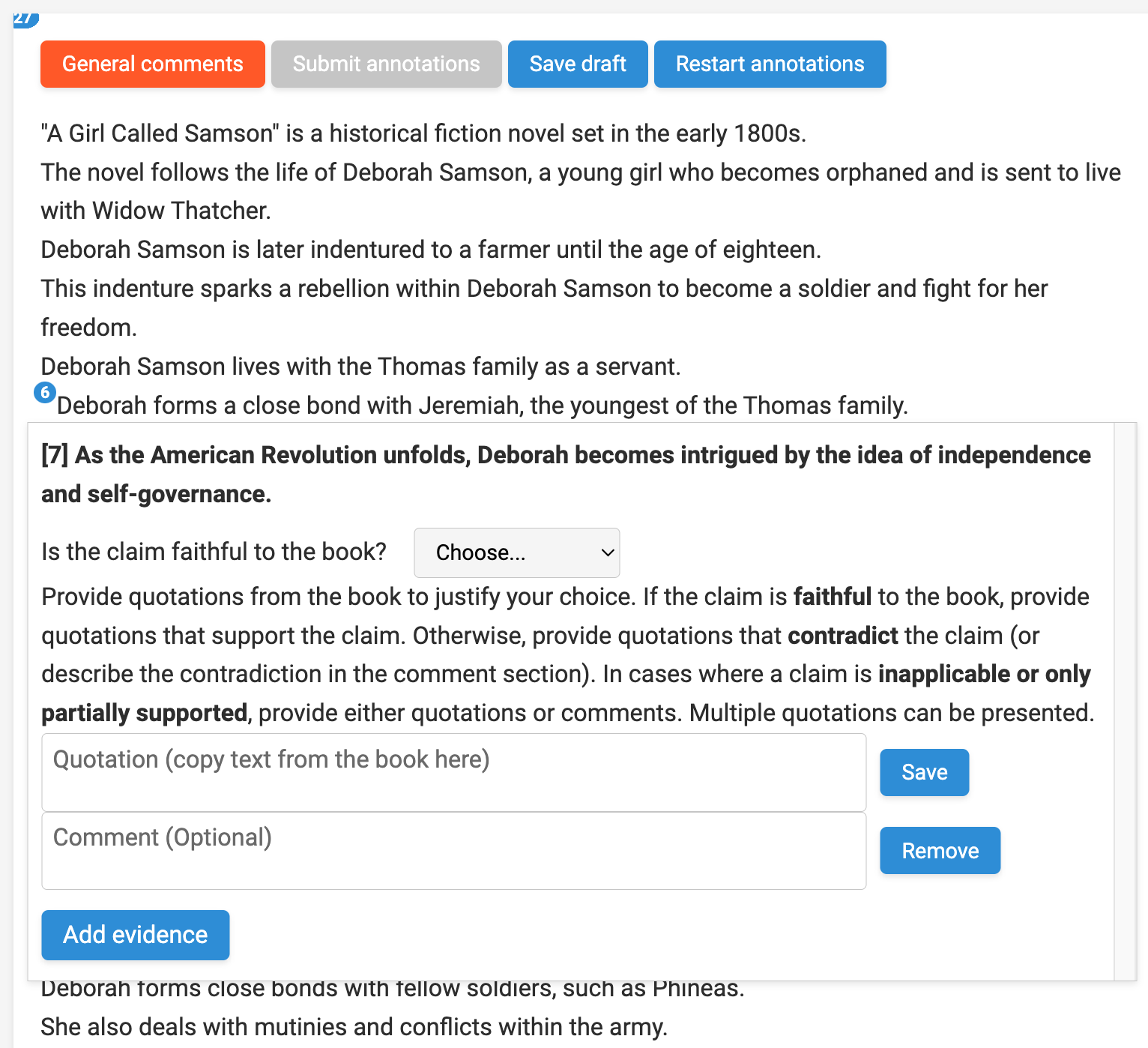}
\caption{Pop-up window showing the interface where the annotators have to select the faithfulness label supplemented by free-form reasoning and evidence extracted from the book.}.
\label{fig:annot_popup}
\end{figure}

\begin{figure}[htbp]
\centering
\includegraphics[width=0.8\linewidth]{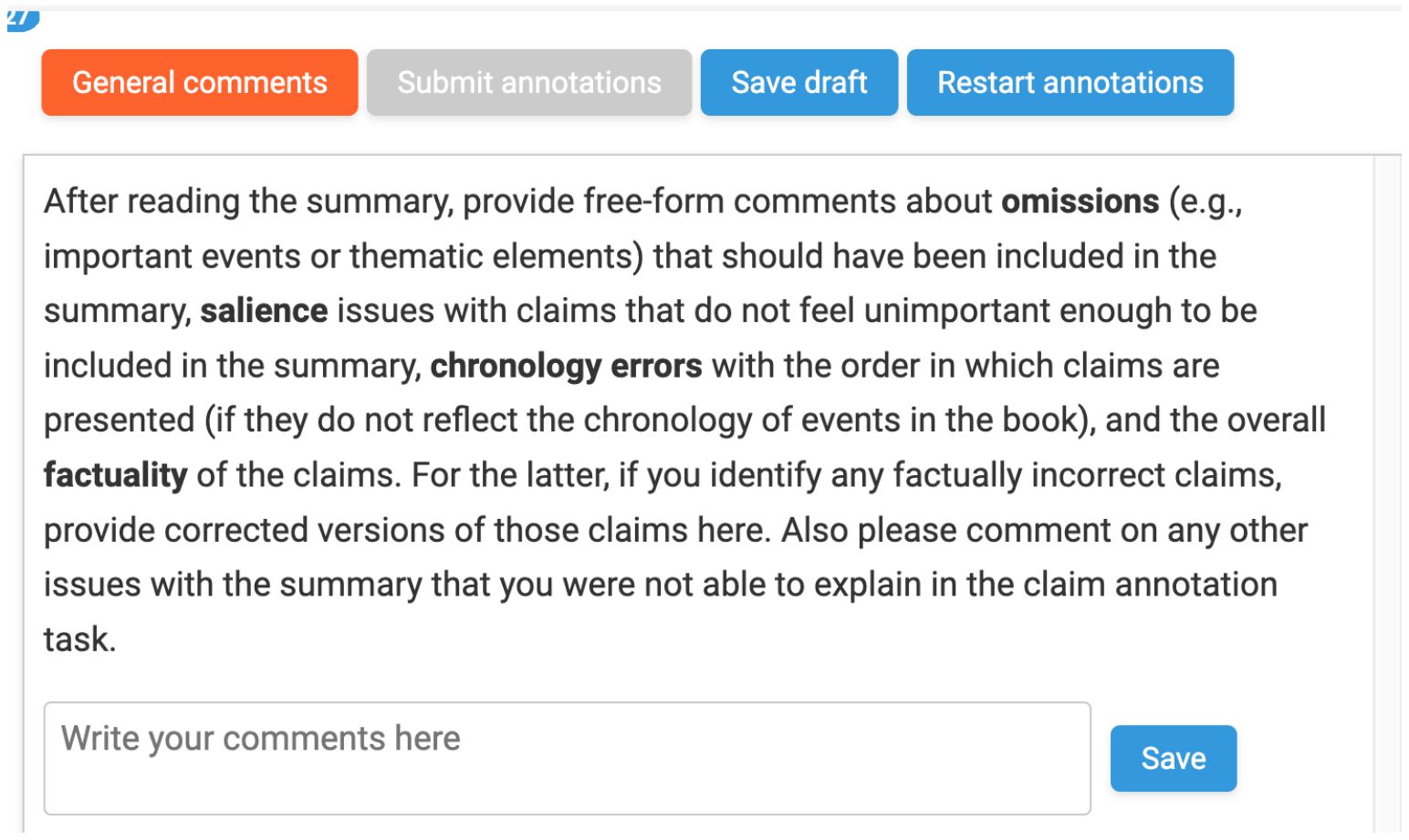}
\caption{Pop-up window prompting the annotator to provide a free-form comment on the quality of summary claims highlighting \textit{omissions}, \textit{salience}, \textit{chronology}, and \textit{factuality} issues.}.
\label{fig:annot_comment}
\end{figure}

In this section, we present details of our annotation task. \autoref{fig:annot_instruction} displays the instructions provided to annotators for evaluating faithfulness. \autoref{fig:annot_interface} illustrates the interface used by annotators for this task. The list of claims is displayed on the left side of the screen, with each claim on a separate line. Content of the book is presented on the right side. Annotators can navigate the book`s content using the scroll function and perform keyword searches to locate relevant information. When annotators hover over a claim, it becomes highlighted, and clicking on it triggers a popup window to appear (see \autoref{fig:annot_popup}).
Given that completing the annotation process takes a considerable amount of time (approximately 1.5h-2.5h), we have implemented a feature that allows annotators to save their work at any point during the annotation process.
Upon completing the annotations, the annotator is required to provide a comment on the overall quality of the summary claims by clicking on \textit{general comments} (see \autoref{fig:annot_comment}).

\paragraph{How do annotators perceive the task?}\label{para:perceive} Annotators highlighted several challenges in assessing the summaries, particularly when dealing with broad claims about themes rather than specific plot points, making it difficult to find relevant supporting evidence within the text. Abstract concepts, like emotions or thematic claims, posed significant obstacles, with some annotators struggling to locate quotations that precisely supported or refuted these claims. They also pointed out the difficulty of evaluating claims that were only partially true, which required more detailed support (see \autoref{tab:annotators-comments-on-task} for actual comments).

% instructions for annotators: fig:annot_instruction
% annot interface: fig:annot_interface
% fig:annot_popup
% fig:annot_comment

\paragraph{Quality of Annotations} 
\label{app:quality_annotation}
We perform two additional analysis experiments that demonstrate the high quality of our dataset: (1) self-consistency of annotations (i.e., how often a single annotator assigns the same label to claims with the same semantic content generated by different models), and (2) inter-annotator agreement on a subset of claims where we had access to another annotator who also read the book.

\begin{itemize}
    \item \textbf{Inter-annotator agreement}: For two books in our dataset, we hired an additional annotator who had also read them to provide overlapping annotations. This resulted in 115 claims with overlapping annotations, allowing us to evaluate the agreement rate between the original and new annotators. The new annotator is 91.30\%, with Cohen’s Kappa of 0.621 (p < .0001), indicating substantial agreement. Unfortunately, annotating the entire dataset with multiple annotators is unfeasible due to the difficulty and high cost of finding multiple individuals who have read the same book. Each annotation costs approximately \$200 to \$250 per book and requires around 10 hours of work.

\item \textbf{Self-consistency}: For each book, an annotator analyzed five summaries, each generated by a different model. To assess self-consistency (intra-annotator agreement), we randomly selected five books and compared the annotations made on the first and last summaries (as per annotation order) for claims with the same semantic content. For example, "Aurora suffers emotional discomfort due to her father's disinterest and her parents' failed marriage" and "Aurora struggles with her father's lack of attention and affection" are semantically equivalent claims from summaries of Wildfire generated by GPT-4 and Claude3, respectively. By comparing the first and last summaries, we evaluated the annotators’ consistency in handling claims after significant time intervals, during which they annotated three additional summaries. Consistency in labels for similar claims across these two summaries would indicate stable judgment and suggest that labels were not arbitrarily assigned.
Out of 127 claims examined in the first summary, 46 had semantically equivalent claims in the last summary, and we found that all 46 of these claims were consistently labeled.
\end{itemize}

\begin{table}[htbp]
    \centering
    \scalebox{0.9}{
    \begin{tabular}{l|cccc}
    \toprule
    Model   & \Faithful & \Unfaithful & \textcolor{violet}{\textit{Partial Support}}  & Can't verify \\
    \midrule
    \chatgpt & 432 & 68 & 79 & 25 \\
   \gpt & 534 & 31 & 108 & 9 \\
         \mixtral & 491 & 83 & 122 & 19 \\
    \gptturbo & 437 & 43 & 68 & 15 \\
    \claude & 528 & 23 & 40 & 3 \\
    \bottomrule
    \end{tabular}
    }
    \caption{Distribution of collected labels by model.}
    \label{tab:human_annotations_results_appendix}
\end{table}

\section{Results of Human Evaluation}
\label{app:additional_by_book_factuality_count}

\begin{figure}[htbp]
\centering
\includegraphics[width=1\linewidth]{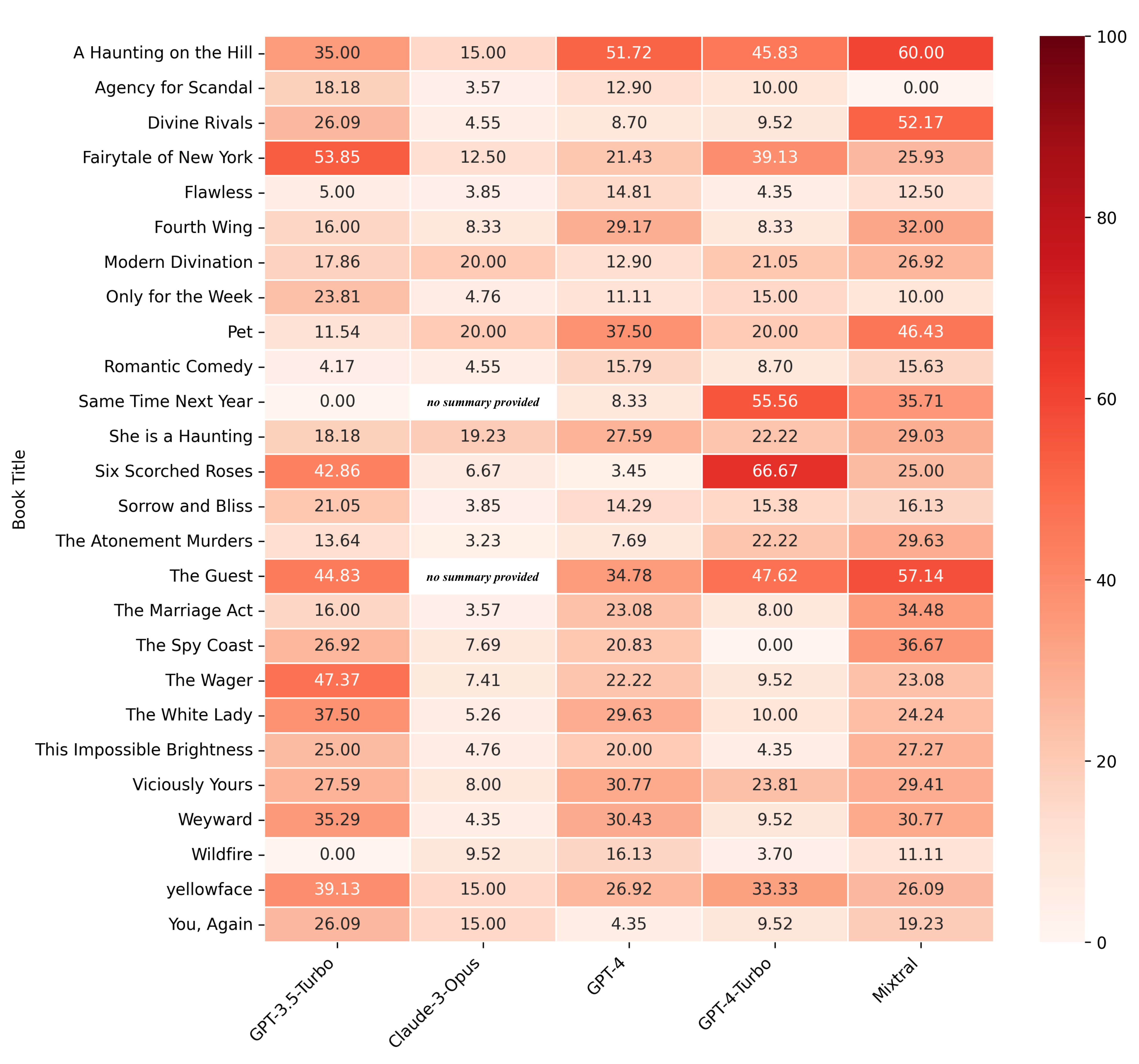}
\caption{Percentage of claims rated \Unfaithful\ or \textcolor{violet}{\textit{Partially Supported}} across models, analyzed by book. \textbf{Only four (4) out of 130 summaries were 100\% \Faithful.} In two cases, \claude\ declined to merge two summaries due to significant content discrepancies (``Same Time Next Year'' and ``The Guest'').}
\label{heatmap:summaries_fact_affected}
\end{figure}

This section provides details on the number of \Unfaithful\ and \textcolor{violet}{\textit{Partially Supported}} claims per summary.   \autoref{heatmap:summaries_fact_affected} presents the percentage of problematic claims (either \Unfaithful\ or \textcolor{violet}{\textit{Partially Supported}}) identified within each model's summaries. Notably, only four (4) out of 130 summaries were rated 100\% \Faithful\ (two by \chatgpt, one by \gptturbo, and one by \mixtral). The remaining summaries varied in accuracy, with some containing up to 66.67\% incorrect or partially incorrect claims.

\section{Analysis of Faithfulness Annotations}
\label{app:criteria}

In this section, we provide additional details on our analysis of faithfulness annotations involving unfaithful claims. Refer to \autoref{tab:faith-anno-criteria} for our general labeling scheme and examples for each category. \autoref{tab:faith-anno-claim-reason} shows the reasoning type distribution for each claim type.

\paragraph{Evidence coverage and reasoning-claim relationship} To investigate the quality of evidence provided by annotators, we analyze the coverage of evidence with respect to the annotators' reasoning. In addition, we also analyze the relationship between the claim and the annotators' reasoning. Results are summarized in \autoref{tab:faith-anno-coverage-rel}. 51.6\% of the time, annotators provide some evidence to justify every component of their reasoning (i.e., \textit{complete coverage}). In 56\% of \textit{partial coverage} (i.e., some part of the reasoning does not have corresponding evidence) cases and all cases of \textit{N/A coverage} (i.e., no evidence is provided at all), the missing evidence is due to the annotator's inability to find any relevant information that either supports or refutes the claim. Qualitatively, for all matched reasoning-evidence pairs, we find that the evidence often does not provide enough context that would allow someone who has not read the book to determine the faithfulness of the claim. As a result of decontextualization, claims always refer to people by name, but evidence often use pronouns instead. The annotator would need to quote a much larger chunk from the book in order for the evidence to include names as well. An even trickier case is that when dealing with high-level claims like ``X is the protagonist of the story" or ``The themes of the story are X, Y, and Z," one needs knowledge of the entire book, but citing the entire book as evidence is trivial. If annotators were to collect self-contained and sufficient evidence for every claim, the task would become significantly more challenging, sometimes even impossible. This difficulty with evidence gathering sheds light on why automatic evaluation does not work so well for this task.

\paragraph{Model-wise analysis} We report model-wise results on reasoning type and reasoning-claim relationship in \autoref{tab:faith-anno-reason-modelwise} and \autoref{tab:faith-anno-rela-modelwise}.

% criteria faithfulness
\begin{table}[htbp]
\centering
\footnotesize
\setlength{\tabcolsep}{5pt}
\renewcommand{\arraystretch}{1.25}
\scalebox{0.95}{
\begin{tabular}{p{0.12\textwidth}p{0.35\textwidth}p{0.48\textwidth}}
\toprule
\multicolumn{1}{c}{\textsc{Label}} & \multicolumn{1}{c}{\textsc{Definition}} & \multicolumn{1}{c}{\textsc{Example ({\color{purple}Claim} // {\color{teal}Reason}})} \\
\hline
\addlinespace
\multicolumn{3}{c}{\includegraphics[height=1.1em]{figures/claim-icon.jpeg}\textbf{ Claim type}} \\
\addlinespace
\hline
Event & Concrete event where someone does something, something happens to someone, etc. & \textit{{\color{purple}Maggie reunites with her old friends and fellow retired spies.} // {\color{teal}Maggie does not reunite with these people.}} \\
\hline
Introspection & Characters' thoughts, feelings, opinions, etc. & \textit{{\color{purple}Justine feels guilty about Amy's death and is haunted by the idea that Amy might be watching her.} // {\color{teal} Justine doesn't feel guilty.}} \\
\hline
\multirow{2}{0.12\textwidth}{Cause/effect} & Goals, motivation, or purposes & \textit{{\color{purple}Charlie Brown decides to return to New York to confront Harry Taylor and pursue a connection with Pete Makris after discovering Harry's infidelity.} // {\color{teal}He is not there to confront Harry.}} \\
\cmidrule(r){2-3}
& Causes or effects of events, actions, thoughts, etc. & \textit{{\color{purple}The discovery of the love story sparks Jade's curiosity about the house and its past inhabitants.} // {\color{teal}Jade's curiosity is not sparked by the love story, but by a dream she had.}} \\
\hline
\multirow{4}{0.12\textwidth}{State} & Relationship between characters & \textit{{\color{purple}Maggie reunites with her old friends and fellow retired spies.} // {\color{teal}Maggie does not reunite with these people.}}\\
\cmidrule(r){2-3}
& Traits of a character & \textit{{\color{purple}The magic of royal fae in ``Viciously Yours" manifests after twenty-five years.} // {\color{teal}It does not manifest after 25 years but becomes full strength at 25 years. They are born with magic.}} \\
\cmidrule(r){2-3}
& State of a character, place, etc. & \textit{{\color{purple} Phillip Hardwicke, a wealthy businessman who was believed to be dead, is revealed to be alive in the story.} // {\color{teal}Bella Hardwicke is revealed to be alive, not Phillip.}} \\
\hline
\multirow{4}{0.12\textwidth}{High-level} & Characteristics of the narrative & \textit{{\color{purple}The narrative style of the book is non-linear and features flashbacks and switches between alternate worlds or viewpoints.} // {\color{teal}The book is almost exclusively from Aurelia's point of view and is linear.}} \\
\cmidrule(r){2-3}
& General story setting & \textit{{\color{purple}The narrative style of the book is non-linear and features flashbacks and switches between alternate worlds or viewpoints.} // {\color{teal}It's set in Adcova, Nyaxia is the name of the goddess.}} \\
\cmidrule(r){2-3}
& Themes & \textit{{\color{purple}The narrative of ``The Guest" explores themes of memory, identity, and the pursuit of understanding within human relationships.} // {\color{teal}It's set in Adcova, Nyaxia is the name of the goddess.}} \\
\hline
\addlinespace
\multicolumn{3}{c}{\includegraphics[height=1.1em]{figures/reasonin-icon.png}\textbf{ Reasoning type}} \\
\addlinespace
\hline
Direct & Reasoning requires only one hop & \textit{{\color{purple}Alex attends a gathering at Victor's house.} // {\color{teal}The book states that the gathering is in Helen's house.}}\\
\hline
\multirow{2}{0.12\textwidth}{Indirect} & Reasoning requires more than one hop & \textit{{\color{purple}Alex and Jack bond over their shared experiences.} // {\color{teal}They don't have any shared experiences, Jack is from a wealthy, privileged home, and while we aren't told much about Alex's background, we know she doesn't live a cosseted life like him.}}\\
\cmidrule(r){2-3}
& Annotator is arguing for a lack of support & \textit{{\color{purple}Maggie is portrayed as a skilled assassin in addition to being a former intelligence officer.} // {\color{teal}No information in the book really supports that.}}\\
\hline
Subjective & Requires subjective judgment & \textit{{\color{purple}Forest is torn between his desire to protect Iris and confronting his past actions.} // {\color{teal}I don't think Forest makes any real effort to confront his past actions, his main motivation is protecting Iris.}}\\
\hline
Extra info & Requires extra/meta information & \textit{{\color{purple}The book ``Wildfire" is the first in the Icebreaker series.} // {\color{teal}No evidence in the book, but this is the second in the series, after ``Icebreaker".}}\\
\bottomrule
\end{tabular}}
\caption{General scheme for assigning labels in our faithfulness annotation analysis along with more examples. This table complements \autoref{tab:faith-anno-results}.}
\label{tab:faith-anno-criteria}
\end{table}

% evidence - reasoning
\begin{table}[htbp]
\centering
\begin{tabular}{lclc}
\toprule
\multicolumn{2}{c}{Evidence Coverage} & \multicolumn{2}{c}{Reasoning-Claim Rel.} \\
\cmidrule(r){1-2} \cmidrule(l){3-4}
Type & Freq & Type & Freq \\
\midrule
Complete & 56.1 & Direct cont. & 50.5 \\
Partial & 34.7 & Indirect cont. & 30.1 \\
Irrel. & 1.5 & Lack of support & 19.4 \\
N/A & 7.7 & & \\
\bottomrule
\end{tabular}
\caption{Results from our analysis on evidence coverage and reasoning-claim relationship.}
\label{tab:faith-anno-coverage-rel}
\end{table}

% tab:faith-anno-claim-reason
\begin{table}[htbp]
\renewcommand{\arraystretch}{1.5}
\centering
\scalebox{0.75}{
\begin{tabular}{l|ccccc}
Claim type & Total count & Direct evidence & Logical inference & Subjective interpretation & Requires meta info \\
\hline
Event & 101 & 46.5 & 46.5 & 3 & 4\\ 
Thought & 26 & 42.3 & 50 & 7.7 & 0\\ 
Cause/effect & 36 & 30.6 & 61.1 & 5.6 & 2.8\\
State & 127 & 39.4 & 48.8 & 5.5 & 6.3\\
High-level & 36 & 16.7 & 55.6 & 16.7 & 11.1\\ 
\hline
\end{tabular}}
\caption{Distribution of reasoning type for each claim type. Apart from total count, all numbers are reported as a percentage.}
\label{tab:faith-anno-claim-reason}
\end{table}

% tab:faith-anno-reason-modelwise
\begin{table}[htbp]
\renewcommand{\arraystretch}{1.5}
\centering
% \small
\scalebox{0.75}{
\begin{tabular}{l|ccccc}
Model & Total count & Direct evidence & Logical inference & Subjective interpretation & Requires meta info \\
\hline
\claude & 12 & 66.7 & 25 & 0 & 8.3 \\ 
\gpt & 26 & 20 & 68 & 12 & 0 \\ 
\gptturbo & 48 & 27.5 & 62.7 & 3.9 & 5.9 \\ 
\chatgpt & 63 & 31.2 & 51.6 & 10.9 & 6.2 \\ 
\mixtral & 76 & 38.5 & 48.7 & 6.4 & 6.4 \\
\hline
\end{tabular}}
\caption{Distribution of reasoning type for different models. Apart from total count, all numbers are reported as a percentage.}
\label{tab:faith-anno-reason-modelwise}
\end{table}

%tab:faith-anno-rela-modelwise
\begin{table}[htbp]
\renewcommand{\arraystretch}{1.5}
\centering
\scalebox{0.85}{
\begin{tabular}{l|cccc}
Model & Total count & Direct contradiction & Indirect contradiction & Lack of support \\
\hline
\claude & 12 & 66.7 & 33.3 & 0 \\ 
\gpt & 26 & 48.3 & 27.6 & 24.1 \\ 
\gptturbo & 48 & 24.5 & 34.7 & 40.8 \\ 
\chatgpt & 63 & 44.1 & 30.9 & 25 \\ 
\mixtral & 76 & 57.7 & 19.2 & 23.1 \\
\hline
\end{tabular}}
\caption{Distribution of reasoning-claim relationship for different models. Apart from total count, all numbers are reported as a percentage.}
\label{tab:faith-anno-rela-modelwise}
\end{table}

%%%%%%%%% COMMENTS ON SUMMARIES %%%%%%%%%%%%%%%%%%%%%%%%%%%%%%%%%%%%
\section{Comment Analysis}
\label{app:comments-section}

\begin{figure}[htbp]
\centering
\includegraphics[width=1\linewidth]{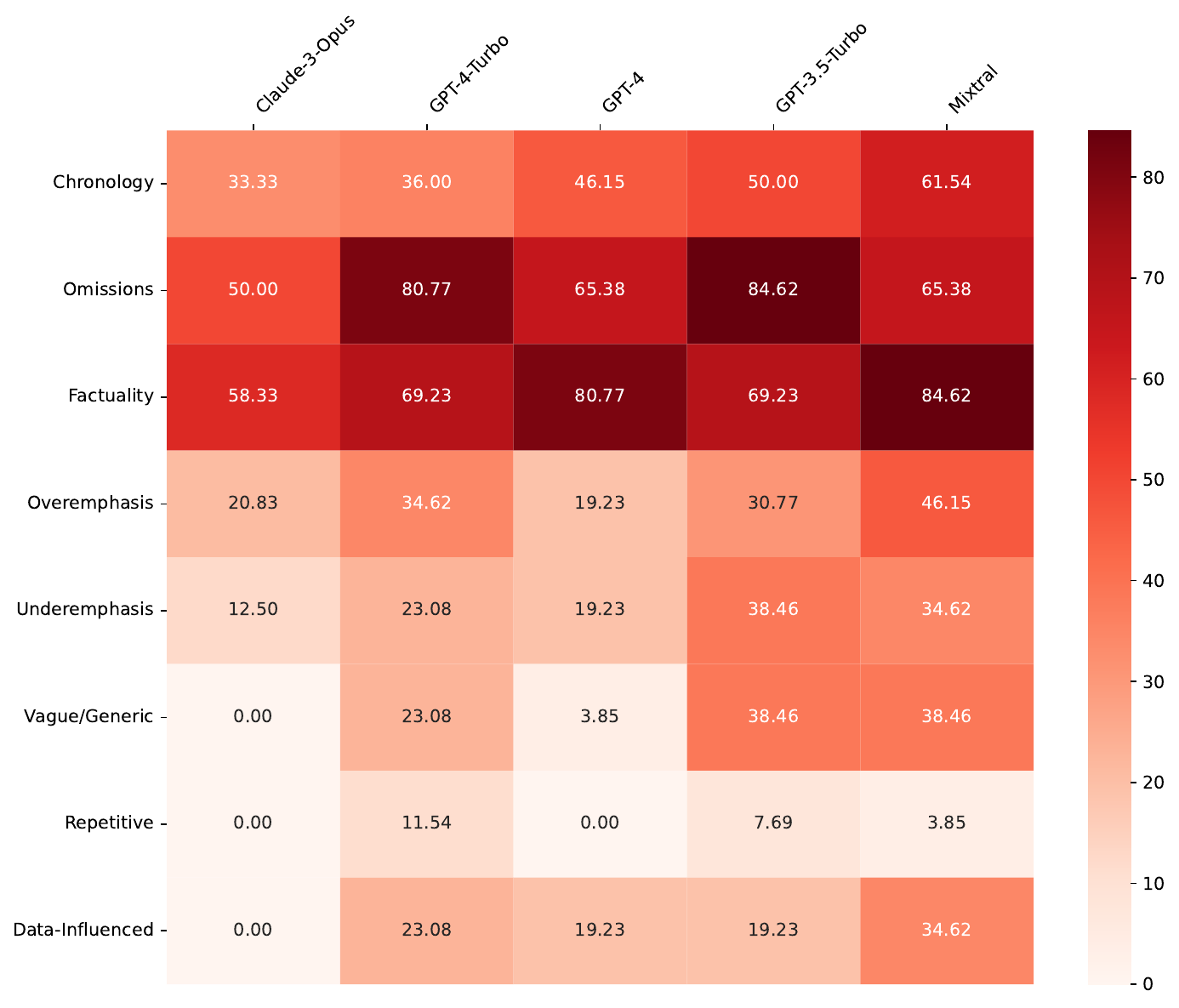}
\caption{Percentage summaries affected by specific issue mentioned in comments by model.}
\label{fig:comments_errors_type_heatmap}
\end{figure}

In this section, we provide additional details regarding our analysis of the comments provided by annotators on the summary claims. \autoref{tab:comments_for_models_examples} features examples of such comments. These comments were further annotated based on the criteria outlined in \autoref{tab:annotator-comments-criteria} and \autoref{tab:omissions-types-criteria}. The distribution of errors is depicted in \autoref{fig:comments_errors_type_heatmap} and \autoref{tab:type_of_omission_errors}.

\autoref{tab:summary-extra-info} displays examples where the models` generation was influenced by information in the front and back matter. \autoref{tab:focus-on-the-end} highlights comments indicating that models may sometimes overly focus on the latter parts of the stories. Lastly, \autoref{tab:annotators-comments-on-task} shares annotators` feedback on the annotation task.

\paragraph{Impact of front and back matter on the summary quality} Books frequently contain additional information beyond the main narrative, including the author`s biography, table of contents, dedications, and more, positioned at the beginning or the end of the book. Ideally, models should exclude this extraneous content, focusing solely on summarizing the core story. However, we have noted that models are sometimes unduly influenced by these elements, which can dominate a significant part of the summary and occasionally compromise its accuracy. Overall, between 19.23\% (\chatgpt\ and \gpt) and 34.62\% (\mixtral) of summaries were affected by such content, either through focusing on this information,\footnote{\textit{``This summary includes a description of who the author thanks at the end of the book which is not important to the plot of the book.''}} confusing story characters with names found in the front and/or back matter,\footnote{\textit{``Clair is not a character in this book. The comments are factual, but of Charlie not Clair.''}} or making up entire narrative based on a single mention.\footnote{\textit{``...claims are very focused on the idea of themes of digital age and the story doesn't cover that at all. Its not even based on a modern world.''} -- author's social media accounts are mentioned at the very end of the book.} \claude\ was the only model seemingly unaffected by the additional information. However, when faced with two summaries—where one primarily summarized the content of the back matter, since it represented the final chunk—the model declined to perform the task. We regard this cautious approach as preferable to introducing unfounded details or irrelevant content.
Examples of such cases are shown in \autoref{tab:summary-extra-info}.

% comment examples: tab:comments_for_models_examples
\begin{table}[!ht]
\centering
\small % 
\scalebox{0.95}{
\begin{tabular}{>{\raggedright\arraybackslash}p{1cm} >{\raggedright\arraybackslash}p{6.2cm} >{\raggedright\arraybackslash}p{6.2cm}}
\toprule
\textbf{Model} & \textbf{\textit{Leaning} {\color{darkgreen}Positive}} & \textbf{\textit{Leaning} {\color{red}Negative}} \\
\midrule
\addlinespace
\multirow{9}{*}{\rotatebox{90}\claude} & \textit{This is an excellently structured summary. It communicates the main plot of the book in a succinct, organised manner, touching on all the pivotal moments in a logical sequence. There is a balance between story and commentary.}  & \textit{Amelia's last name isn't very relevant to the summary. She is referred to by it once in the book. Taking out her last name would be more effective. Rennick's mother's death is also omitted which is a key point in the story. The other missing key point is that Amelia is believed to be a human. That has major implications and sets the stage for the events in the story. Overall this summary compared the first was a much better summary of the story because it contains key points and not generalizations.} \\
\addlinespace
\multirow{8}{*}{\rotatebox{90}\gptturbo} & \textit{Overall, I think this is a good summary. It hits the major points of the book and the three stories are each in order. However, the book jumps back and forth between the three stories and this summary goes through each one separately. It also leaves out Violet drinking a tea that causes an abortion which is an important part of her character's.} & \textit{This is not a strong summary of the book. The entire section about visiting and living with Jerry is missing. There is no mention of the paparazzi pics being leaked. This summary seems to focus more on the message of the book rather than the plot. Many of the claims seems to be paraphrased versions of each other and keep repeating the same ideas of this book being about overcoming challenges, facing insecurities etc.} \\
\addlinespace
\multirow{10}{*}{\rotatebox{90}\gpt}      & \textit{This summary is written in chronological order, it accurately categorizes the excerpt of Wreck the Halls and the claims mentioned here are either True or Partially True. It is easy to follow and is not repetitive. This is one of the stronger summaries. The claims were easy to verify and with the partially true claims - there were only minor issues. However, one aspect that is ignored in all the summaries including this one is the importance of Sumner’s family. It touches on his view of his parents as a couple and his relationship with his grandmother. (...)} & \textit{The majority of these claims were true, but the overall summary does a poor job of following the plot. It skips over many important plot points and does a poor job of describing the main characters. Similar to another one of the summaries, this one makes a lot of broad thematic claims but misses the overall storyline. If I hadn't read the book, I would not understand this summary.} \\
\addlinespace
\multirow{10}{*}{\rotatebox{90}\chatgpt} & \textit{This is a good summary with most of the plot points covered. The chronological sequence is largely appropriate. Though I would move the claim about Sally and Noah's marriage closer to the conclusion as it happens at the end of the book. In one claim, Jerry is listed as he uncle, though he is her stepfather. I think there should also be a claim that Sally goes to live at Noah's place during the pandemic, as they get closer from their email exchanges. Otherwise, the summary communicates the heart of the book.} & \textit{This summary has a lot of issues including chronology problems, irrelevant information, and factual errors. Multiple claims draw from the acknowledgments and about the author section and these claims appear in the middle of the summary. Factual errors include that it is an assassin, not Diana that murders Gavin in Bangkok, Declan and Ingrid not being involved in past crimes in Malta, and Luther and Callie not having connections to Maggie's past. This summary misses some important points including that Maggie's husband was killed as a result of the Malta/ Cyrano mission.} \\
\addlinespace
% \addlinespace
\multirow{10}{*}{\rotatebox{90}\mixtral}        & \textit{The summary focuses on most of the key points mentioned in the story. It starts in chronological order and focuses on the evolvement of their relationship from a business arrangement towards a happily ever after. It describes the initial and final scenes well, including all relevant details.  However, it misses certain events such as getting to know each other, the green card interview, first kiss (...).} & \textit{This is not a good summary and should not be used. there are too many false statements. Also, the jumping back and forth in the narrative makes it hard to follow.} \\
\addlinespace
\bottomrule
\end{tabular}
}
\caption{Examples of positive and negative comments submitted by the annotators for specific models}
\label{tab:comments_for_models_examples}
\end{table}

% annot criteria: tab:annotator-comments-criteria
\begin{table}[ht]
   \centering
   \small
   \scalebox{0.85}{
   \begin{tabular}{lp{0.4\textwidth}p{0.4\textwidth}}
       \toprule
       \textbf{Issue} & \textbf{Definition} & \textbf{Example} \\
       \midrule
       \addlinespace
       \textsc{Chronology} & Issues with the chronological ordering of claims. & \textit{ (...) though it has some chronology problems (Ari's proposal comes after the run in the city, which comes after Josh and Radhya open the pop-up) (...)} \\
       \addlinespace
       \textsc{Omissions} & The annotator mentions any omissions of content that should have been included in the summary. & \textit{Omissions: Dr. Rob Valentine groomed Summer from when she was a teenager until she turned 18 and then started an intimate relationship. He later leaves Summer and marries her sister. Even after he is married, he doesn't let Summer move on.}
       \\
       \addlinespace
       \textsc{Factuality} & Issues with factuality are explicitly mentioned by the annotator. Note that this category correlates partially with the annotated factuality errors.  & 
       \textit{There were some serious issues with this summary. The first being that the book is referred to as "The Retirement Plan" twice in the summary which is the incorrect title.} \\
       \addlinespace
       \textsc{Overemphasis} & Too much emphasis put on less significant events or characters. & \textit{Salience: Charles is not an important character, he is the manager of the guest house where she stays in New York, and she only chats to him a couple of times.} \\
       \addlinespace
       \textsc{Underemphasis} & Certain events or characters are mentioned but too little emphasis is put on their importance for the story. & \textit{There is not enough emphasis on the relationship between Justine and Dom, who later becomes her husband.} \\
       \addlinespace
       \textsc{Vague/Generic} & Vague or generic claims included in the summary. & \textit{Most of the sentences at the end of the summary are generalized and there are no substantial facts.}\\
       \addlinespace
       \textsc{Repetitive} & Repetitive claims included in the summary. & \textit{As i noted in the annotations, claims 16 and 17 were repetitious and not necessary.} \\
       \addlinespace
       \textsc{Data-influenced} & The summary was influenced by front and/or back matter. & [n/a: judgment for this category was made by one of the co-authors during analysis] \\
       \addlinespace
       \midrule
       \textsc{Comprehensive} & The annotator praises the summary for being comprehensive. & \textit{Out of all the summary claims, this feels the most relevant and comprehensive of the key events that take place.} \\
       \addlinespace
       \textsc{Well-done} & The annotator praises the summary for being well-done. & \textit{This was good, things were in sequence, and the main points were covered.} \\
       \bottomrule
   \end{tabular}
   }
   \caption{Categories used for the analysis of annotators' comments on the quality of the entire summary.}
   \label{tab:annotator-comments-criteria}
\end{table}

% omissions: tab:omissions-types-criteria
\begin{table}[ht]
   \centering
   \small
   % \scalebox{0.9}{
   \begin{tabular}{lp{0.35\textwidth}p{0.4\textwidth}}
       \toprule
       \textbf{Omission Type} & \textbf{Definition} & \textbf{Example} \\
       \midrule
       \textsc{Characters} & Summary fails to mention important characters. & \textit{This summary excluded a lot of the main plot points (...) and the very important principal antagonists Alma, Thomas, and Marion. Alma and Thomas are present-day representations of colonization and investors in the project to turn Nhà Hoa into a bed-and-breakfast. } \\
       \addlinespace
       \textsc{Events} & Summary fails to mention important events and/or turning points.  & 
       \textit{Some important events in the book were omitted, such as the part where Alex follows a group of young people to a house and has sex with a girl's boyfriend, the part where she sneaks into a club and pretends to be a little boy's nanny, and the part where she follows Margaret to her home. These events are filled with tension, showcase Alex's daring exploits, add a deeper layer of meaning to the story, and ultimately propel the narrative, so they should have been included. Alex almost drowns in the beginning of the novel, a frightening incident that she mentions to Victor during the party.} \\
       \addlinespace
       \textsc{Attributes} & Summary fails to mention attributes -- details about the events, interactions, characters, etc. & \textit{(...) but the overall summary misses a huge plot point of Carver and Roman being the same person.} \\
       \addlinespace
       \textsc{Relations} & Summary fails to mention specific relationships between the characters which are important to the plot. & \textit{There is no mention of Clover and Amos the Desert King being mates. Nor Fawn and Dean the Garden King being mates.} \\
       \addlinespace
       \textsc{Themes} & Summary fails to depict main/important themes of the novel. & \textit{Water, pools, and beaches are recurring metaphors in the book, yet they aren't mentioned or highlighted in any of the claims.} \\
      % \addlinespace
       \bottomrule
   \end{tabular}
   \caption{Description of omission categories used for annotating comments provided by our evaluators. Omissions were annotated in two steps: (1) a binary choice (either omissions were mentioned or not), and (2) categorization.}
   % }
   \label{tab:omissions-types-criteria}
\end{table}

% issues in comments heatmap: fig:comments_errors_type_heatmap

% omission errors: tab:type_of_omission_errors
\begin{table}[ht]
\centering
\scalebox{0.9}{
\begin{tabular}{lccccc}
\toprule
\textsc{Omission Type} & \claude & \gptturbo & \gpt & \chatgpt & \mixtral \\
\midrule
Characters & 4.17 & 23.08 & 7.69 & 7.69 & 23.08 \\ 
Events & 33.33 & 57.69 & 38.46 & 65.38 & 38.46 \\
Attributes & 16.67 & 38.46 & 34.62 & 34.62 & 38.46 \\
Relations & 8.33 & 11.54 & 11.54 & 11.54 & 15.38 \\
Theme & 12.50 & 15.38 & 3.85 & 15.38 & 19.23 \\
\bottomrule
\end{tabular}
}
\caption{Percentage of summaries affected by specific type of omission error by model.}
\label{tab:type_of_omission_errors}
\end{table}

% extra info in summaries: tab:summary-extra-info
\begin{table}[ht]
    \centering
    \footnotesize
    \scalebox{0.88}{
    \begin{tabular}{lp{0.4\textwidth}p{0.55\textwidth}}
        \toprule
        \textbf{Model} & \textbf{Comment} & \textbf{Summary} \\
        \midrule
        \addlinespace
        \multirow{15}{*}{\rotatebox{90}\claude} & \textit{None of the claims is completely factually correct, or focused on the book as they ought to be. The first two claims are the only ones where the protagonist Alex is mentioned, but both claims are still wrong. The other claims are incomprehensible, have absolutely nothing to do with the book, and are therefore inadmissible. In fact, the claims appear to be focused on a different topic entirely, not the book. Please note that the quotes I included below don't necessarily prove any of the claims wrong (or right, for that matter). (...)} 
        \newline {\color{purple} \textbf{NOTE:} The model declines to merge two summaries due to significant discrepancies in content.} & \textit{I apologize for the confusion, but the provided summaries do not appear to be related to the same story. Summary 1 describes a novel about a young woman named Alex and her experiences in the Hamptons, while Summary 2 simply states that the provided excerpt is a table of contents or metadata without any actual story content. As there is no coherent narrative between the two summaries, it is not possible to merge them into a single, comprehensive summary. In order to create a merged summary, I would need access to summaries that are all related to the same story, with each summary focusing on different parts or aspects of that story. If you are able to provide a series of summaries that are all connected to the same narrative, I would be happy to assist in combining them into a single, cohesive summary that includes key events, backgrounds, settings, characters, objectives, and motivations, while also introducing major elements and organizing the summary in a chronological manner. Please let me know if you have any other summaries related to the story mentioned in Summary 1, as that would allow me to proceed with the merging process you have requested.} \\
        \addlinespace
        \multirow{15}{*}{\rotatebox{90}\gpt} & \textit{This summary is largely factual and looks at the overall arc of the text, though it has some chronology problems (Ari's proposal comes after the run in the city, which comes after Josh and Radhya open the pop-up), and focuses disproportionately on the author's conversation with another writer, Kate Robb, as opposed to the text itself. It also doesn't discuss the nonlinear storytelling.} 
         \newline {\color{purple} \textbf{NOTE:} The model includes back matter information. While accurate, it should not have been included in the summary.} & \textit{(...) Goldbeck's insightful writing and clever comedic timing elevate the narrative, making it stand out and resonate with a broad variety of readers. Reflecting upon shared preferences like tequila and various NYC spots, Goldbeck successfully manages to craft relatable characters with everyday experiences, despite their complex emotional journey. The author's creative process in the narrative, including the decision to cut a subplot involving Ari working as a comedian on a cruise ship, adds a layer of interesting behind-the-scenes perspective. The narrative style, which pays homage to classic 2000s romantic comedies, echoes elements of escapism, emphasizing the importance of literature during unprecedented times like the global pandemic. (...)} \\
        \addlinespace
        \multirow{15}{*}{\rotatebox{90}\gptturbo} & \textit{This set of summary claims are very focused on the idea of themes of digital age and the story doesn't cover that at all. Its not even based on a modern world. Chronology wasn't relevant as majority of the claims weren't correct, applicable or verifiable. Factuality was on zero and the claims don't cover any of the key events that occurs except that there was themes of life and death.} 
         \newline {\color{purple} \textbf{NOTE:} The model hallucinates information based on brief mentions of author's social media accounts. }
        & \textit{(...) Parallel to Lilith and Vale's journey, the story delves into the lives of modern city-dwellers: Alex, a budding Instagram influencer; Jordan, a TikTok content creator; Sam, a Twitter-using political activist; and Taylor, who leverages Facebook for business and family connections. Each character embodies the diverse ways social media influences the pursuit of fulfillment, showcasing the profound effects of these platforms on personal and collective experiences. (...) Themes of authenticity versus constructed online personas, the search for connectivity amidst digital isolation, and the double-edged sword of viral fame are explored, reflecting on the profound impacts of social media in modern society. (...) In summary, this narrative offers a comprehensive exploration of human experiences, spanning the ethereal and the digital. By paralleling the mystical journey of Lilith against the backdrop of a society deeply influenced by social media, the story invites readers into an immersive world where the boundaries between life and death blur, and where the quest for understanding and connection transcends the mortal coil and the digital divide. Through the intertwined lives of its characters, the story serves as a reflection on the complexities of the digital age, highlighting the profound and varied impacts of social media on the contemporary human condition.} \\
        \addlinespace
        \bottomrule
    \end{tabular}
    }
    \caption{Examples of summaries influenced by front/back matter information along with the annotators` comments. The \claude\ example was excluded from the analysis because the model failed to generate a summary. Although not ideal, this behavior is arguably better than the model fabricating content.}
    \label{tab:summary-extra-info}
\end{table}

% tab:focus-on-the-end
\begin{table}[ht]
\centering
\small
\begin{tabular}{@{}p{0.2\linewidth}p{0.75\linewidth}@{}}
\toprule
 Model & \textsc{Comments} \\ 
\midrule
\addlinespace
\claude\ & \textit{``It also focuses extensively on the last couple chapters of the book. This is the only summary so far that has included claims about the very last chapter of the book that is from Dacre's point of view.''} \\
\addlinespace
\claude\ & \textit{``(...) and hits the main thematic elements of the text, though it disproportionately addresses the epilogue over other portions of the text (...)''} \\
\addlinespace
\claude\ & \textit{``This summary included a lot of true elements, but also included many irrelevant details not integral to the plot. This is especially true for the end of the book.''} \\ 
\addlinespace
\gptturbo\ & \textit{``This summary focuses heavily on the end of the book and misses plot points that happen in the beginning of the book.''} \\ 
\addlinespace
\gpt\ & \textit{``(...) and focuses disproportionately on the author's conversation with another writer, Kate Robb, as opposed to the text itself.''} [the interview is included at the end of the book]\\ 
\addlinespace
\mixtral\ & \textit{``The summary puts an emphasis on Part 4 of the book which is not in proportion to the rest of the book''} [Part 4 is the last part] \\ 
\addlinespace
\bottomrule
\addlinespace
\end{tabular}
\caption{Comments from annotators on models' focus towards the book's end}
\label{tab:focus-on-the-end}
\end{table}

%%%%%%%%%%%%%%%%%%%%%%%%%%%%%%%%%%%%%%%%%%%%%%%%%%%%%%%%%%%%
%%%%%%%%%%%%%%% EXPERIMENTS %%%%%%%%%%%%%%%%%%%%%%%%%%%%%
%%%%%%%%%%%%%%%%%%%%%%%%%%%%%%%%%%%%%%%%%%%%%%%%%%%%%%%%%%%%

\section{Details on Experimental Setup}
\label{app:experiment}

In this section, we provide further details on our experimental setup complemented with further results.

\subsection{Implementation details}
\label{app:implementation_detail}

For BM25-based evidence retrieval, we use the text of \texttt{e-books} purchased from \texttt{amazon.com}, split into passages of up to 256 tokens each. The search is restricted to the book content, and we set $k=5$ to retrieve the top 5 most relevant passages as evidence.

\subsection{Additional Results}
\label{app:additioanl resutls}

Results for each evidence extraction method broken down by summarizer can be found in \autoref{tab:benchmark_results_full}. We also report book-wise precision and recall for each evidence extraction method: (1) No-Context (\autoref{tab:performance_metrics_no_evidence}); (2) BM25 (\autoref{tab:performance_metrics_bM25}); (3) Human evidence (\autoref{tab:performance_metrics_human}); (4) Entire book (\autoref{tab:performance_metrics_WB}). Further results for the entire book (EB) prompting can be found in \S\ref{app:whole-book-approach}.

% different evidence tab:benchmark_results_full
\begin{table}[htbp]
\centering
\scalebox{0.9}{
\renewcommand{\arraystretch}{1.2}
\begin{tabular}{l c c c c c c}
\toprule
 \multirow{2}{*}{\textit{Summarized by}} & \multicolumn{2}{c}{\textbf{No-Context}} & \multicolumn{2}{c}{\textbf{BM25}} &  \multicolumn{2}{c}{\textbf{Human Evidence}} \\
\cmidrule(lr){2-3} \cmidrule(lr){4-5} \cmidrule(lr){6-7} 
  & \textbf{\Faithful}  & \textbf{\Unfaithful}    & \textbf{\Faithful}  & \textbf{\Unfaithful}   & 
  \textbf{\Faithful}  & \textbf{\Unfaithful}  \\
\midrule
\addlinespace
\multicolumn{7}{l}{\textit{F1 score against the human annotations  }  }  \\ \addlinespace
\chatgpt & 0.727 & 0.261 & 0.835 & 0.476 &  0.712 & 0.430\\
\mixtral  & 0.643 & 0.183 & 0.837 & 0.244 &  0.784 & 0.406 \\
\gpt & 0.687 & 0.130 & 0.794 & 0.088 &  0.721 & 0.207 \\
\gptturbo & 0.634 & 0.033 & 0.887 & 0.080  & 0.792 & 0.139 \\
\claude & 0.674 & 0.000 & 0.738 & 0.000  & 0.684 & 0.031 \\
\\
\midrule
Overall  & 0.681 & 0.124 & \textbf{0.826} & 0.215  & 0.755 & \textbf{0.259}  \\

\bottomrule

\addlinespace
\multicolumn{7}{l}{\textit{Token length of the given evidence against prediction label}  }  \\ \addlinespace
\chatgpt & \multicolumn{2}{c}{0.0}  & 1136.4 & 1131.0 & 292.4 & 126.5 \\
\mixtral & \multicolumn{2}{c}{0.0}  & 1139.5 & 1132.9 & 211.7 & 153.4 \\
\gpt & \multicolumn{2}{c}{0.0}  & 1138.6 & 1132.5  & 241.7 & 160.8 \\
\gptturbo & \multicolumn{2}{c}{0.0}  & 1141.9 & 1138.3& 257.8 & 152.9 \\
\claude & \multicolumn{2}{c}{0.0}  & 1134.5 & 1128.6 & 214.4 & 151.5 \\
\midrule
Average & \multicolumn{2}{c}{0.0} & 1138.2 & 1132.6 & 243.6 & 149.0 \\
\bottomrule
\end{tabular}
}
\caption{Comparison of automatic evaluation using \gptturbo\ based on different evidence extraction methods. We also presents the F1 score and token length of the extracted evidence for each summarizer. Overall mean values were calculated using all the claims across \fables.}
\label{tab:benchmark_results_full}
\end{table}

%  no evidence tab:performance_metrics_no_evidence
\begin{table}[ht]
\centering
\scalebox{0.85}{
\begin{tabular}{>{\raggedright\arraybackslash}p{3.5cm} 
                >{\centering\arraybackslash}p{2.3cm} 
                >{\centering\arraybackslash}p{2.3cm} 
                >{\centering\arraybackslash}p{2.3cm} 
                >{\centering\arraybackslash}p{2.3cm}}
\toprule
\textsc{Title} & \textsc{Pr-\textbf{\Faithful}} & \textsc{Re-\textbf{\Faithful}} & \textsc{Pr-\textbf{\Unfaithful}} & \textsc{Re-\textbf{\Unfaithful}} \\\midrule
A Haunting on the Hill                & 0.821                   & 0.329                & 0.230                    & 0.650                 \\
Agency for Scandal                    & 0.960                   & 0.133                & 0.034                    & 0.833                 \\
Divine Rivals                         & 0.960                   & 0.156                & 0.140                    & 0.917                 \\
Fairytale of New York                 & 1.000                   & 0.123                & 0.174                    & 1.000                 \\
Flawless                              & 0.950                   & 0.217                & 0.012                    & 0.500                 \\
Fourth Wing                           & 1.000                   & 0.169                & 0.112                    & 1.000                 \\
Modern Divination                     & 1.000                   & 0.062                & 0.092                    & 1.000                 \\
Only For The Week                     & 0.893                   & 0.186                & 0.056                    & 0.500                 \\
Pet                                   & 0.871                   & 0.151                & 0.121                    & 0.881                 \\
Romantic Comedy                       & 1.000                   & 0.170                & 0.020                    & 1.000                 \\
Same Time Next Year                   & 0.667                   & 0.161                & 0.204                    & 0.700                 \\
She Is a Haunting                     & 0.917                   & 0.220                & 0.065                    & 0.667                 \\
Six Scorched Roses                    & 0.750                   & 0.179                & 0.228                    & 0.701                 \\
Sorrow and Bliss                      & 0.983                   & 0.197                & 0.029                    & 0.750                 \\
The Atonement Murders                 & 1.000                   & 0.067                & 0.069                    & 1.000                 \\
The Guest                             & 0.688                   & 0.182                & 0.253                    & 0.810                 \\
The Marriage Act                      & 1.000                   & 0.101                & 0.041                    & 1.000                 \\
The Spy Coast                         & 0.864                   & 0.151                & 0.103                    & 0.790                 \\
The Wager                             & 1.000                   & 0.495                & 0.085                    & 1.000                 \\
The White Lady                        & 0.750                   & 0.045                & 0.151                    & 0.938                 \\
This Impossible Brightness            & 1.000                   & 0.147                & 0.044                    & 1.000                 \\
Viciously Yours                       & 0.950                   & 0.200                & 0.113                    & 0.833                 \\
Weyward                               & 0.947                   & 0.504                & 0.161                    & 0.750                 \\
Wildfire                              & 1.000                   & 0.229                & 0.036                    & 1.000                 \\
You, Again                            & 1.000                   & 0.214                & 0.041                    & 1.000                 \\
Yellowface                            & 0.933                   & 0.163                & 0.119                    & 0.938                 \\
\bottomrule
\end{tabular}
} % End scalebox
\caption{Precision (\textsc{Pr}) and Recall (\textsc{Re}) from LM evaluation using \gptturbo\ \textbf{no context} for each book.}
\label{tab:performance_metrics_no_evidence}
\end{table}

%  human evidence
\begin{table}[ht]
\centering
\scalebox{0.85}{
\begin{tabular}{>{\raggedright\arraybackslash}p{3.8cm} 
                >{\centering\arraybackslash}p{2.3cm} 
                >{\centering\arraybackslash}p{2.3cm} 
                >{\centering\arraybackslash}p{2.3cm} 
                >{\centering\arraybackslash}p{2.3cm}}
\toprule
\textsc{Title} & \textsc{Pr-\Faithful} & \textsc{Re-\Faithful} & \textsc{Pr-\Unfaithful} & \textsc{Re-\Unfaithful}  \\\midrule
    A Haunting on the Hill             & 0.967                   & 0.517                & 0.337                    & 0.950                 \\
Agency for Scandal                 & 1.000                   & 0.570                & 0.063                    & 1.000                 \\
Divine Rivals                      & 0.980                   & 0.427                & 0.190                    & 0.917                 \\
Fairytale of New York              & 1.000                   & 0.230                & 0.195                    & 1.000                 \\
Flawless                           & 1.000                   & 0.352                & 0.033                    & 1.000                 \\
Fourth Wing                        & 0.985                   & 0.662                & 0.244                    & 0.950                 \\
Modern Divination                  & 1.000                   & 0.409                & 0.138                    & 1.000                 \\
Only For The Week                  & 1.000                   & 0.509                & 0.119                    & 1.000                 \\
Pet                                & 0.981                   & 0.676                & 0.242                    & 0.833                 \\
Romantic Comedy                    & 1.000                   & 0.204                & 0.024                    & 1.000                 \\
Same Time Next Year                & 0.851                   & 0.780                & 0.300                    & 0.600                 \\
She Is a Haunting                  & 0.985                   & 0.641                & 0.186                    & 0.750                 \\
Six Scorched Roses                 & 0.907                   & 0.521                & 0.334                    & 0.759                 \\
Sorrow and Bliss                   & 1.000                   & 0.340                & 0.038                    & 1.000                 \\
The Atonement Murders              & 1.000                   & 0.333                & 0.093                    & 1.000                 \\
The Guest                          & 0.868                   & 0.644                & 0.391                    & 0.589                 \\
The Marriage Act                   & 1.000                   & 0.360                & 0.049                    & 1.000                 \\
The Spy Coast                      & 0.975                   & 0.392                & 0.168                    & 0.933                 \\
The Wager                          & 1.000                   & 0.810                & 0.220                    & 1.000                 \\
The White Lady                     & 1.000                   & 0.311                & 0.201                    & 1.000                 \\
This Impossible Brightness         & 1.000                   & 0.378                & 0.049                    & 1.000                 \\
Viciously Yours                    & 0.921                   & 0.508                & 0.156                    & 0.688                 \\
Weyward                            & 1.000                   & 0.598                & 0.157                    & 1.000                 \\
Wildfire                           & 1.000                   & 0.387                & 0.042                    & 1.000                 \\
You, Again                         & 0.967                   & 0.469                & 0.048                    & 0.667                 \\
Yellowface                         & 0.935                   & 0.461                & 0.153                    & 0.813                 \\
\bottomrule
\end{tabular}
} % End scalebox
\caption{Results of average Precision (\textsc{Pr}) and Recall (\textsc{Re}) estimated by \textbf{human evidence} and LM evaluation using \gptturbo\ for each book.}
\label{tab:performance_metrics_human}
\end{table}

% BM 25 evidence tab:performance_metrics_bM25
\begin{table}[ht]
\centering
\scalebox{0.85}{
\begin{tabular}{>{\raggedright\arraybackslash}p{3.8cm} 
                >{\centering\arraybackslash}p{2.3cm} 
                >{\centering\arraybackslash}p{2.3cm} 
                >{\centering\arraybackslash}p{2.3cm} 
                >{\centering\arraybackslash}p{2.3cm}}
\toprule
\textsc{Title} & \textsc{Pr-\Faithful} & \textsc{Re-\Faithful} & \textsc{Pr-\Unfaithful} & \textsc{Re-\Unfaithful}  \\\midrule
    A Haunting on the Hill & 0.902 & 0.781 & 0.520 & 0.550 \\
    Agency for Scandal & 0.974 & 0.721 & 0.067 & 0.333 \\
    Divine Rivals & 0.907 & 0.541 & 0.209 & 0.556 \\
    Fairytale of New York & 0.874 & 0.712 & 0.263 & 0.556 \\
    Flawless & 1.000 & 0.663 & 0.056 & 1.000 \\
    Fourth Wing & 0.953 & 0.759 & 0.244 & 0.600 \\
    Modern Divination & 0.950 & 0.644 & 0.142 & 0.700 \\
    Only For The Week & 0.967 & 0.789 & 0.167 & 0.417 \\
    Pet & 0.907 & 0.604 & 0.175 & 0.649 \\
    Romantic Comedy & 1.000 & 0.675 & 0.044 & 1.000 \\
    Same Time Next Year & 0.836 & 0.859 & 0.333 & 0.425 \\
    She Is a Haunting & 0.949 & 0.772 & 0.100 & 0.250 \\
    Six Scorched Roses & 0.816 & 0.500 & 0.269 & 0.616 \\
    Sorrow and Bliss & 1.000 & 0.588 & 0.077 & 1.000 \\
    The Atonement Murders & 1.000 & 0.642 & 0.115 & 1.000 \\
    The Guest & 0.845 & 0.737 & 0.395 & 0.598 \\
    The Marriage Act & 0.987 & 0.833 & 0.119 & 0.833 \\
    The Spy Coast & 0.953 & 0.527 & 0.216 & 0.738 \\
    The Wager & 0.958 & 0.862 & 0.100 & 0.167 \\
    The White Lady & 0.836 & 0.628 & 0.097 & 0.250 \\
    This Impossible Brightness & 0.961 & 0.607 & 0.022 & 0.250 \\
    Viciously Yours & 0.929 & 0.725 & 0.172 & 0.521 \\
    Weyward & 0.969 & 0.774 & 0.300 & 0.583 \\
    Wildfire & 0.980 & 0.664 & 0.043 & 0.750 \\
    You, Again & 0.960 & 0.609 & 0.015 & 0.333 \\
    Yellowface & 0.941 & 0.695 & 0.204 & 0.562 \\
\bottomrule
\end{tabular}
} 
\caption{Results of average Precision (\textsc{Pr}) and Recall (\textsc{Re}) estimated by \textbf{BM25} retriever and LM evaluation using \gptturbo\ for each book.}
\label{tab:performance_metrics_bM25}
\end{table}

% prompting with book  tab:performance_metrics_WB
\begin{table}[ht]
\centering
\scalebox{0.80}{
\begin{tabular}{ >{\centering\arraybackslash}p{2.3cm} 
                >{\raggedright\arraybackslash}p{3.8cm} 
                >{\centering\arraybackslash}p{2.3cm} 
                >{\centering\arraybackslash}p{2.3cm} 
                >{\centering\arraybackslash}p{2.3cm} 
                >{\centering\arraybackslash}p{2.3cm}}
\toprule
Evaluation LM & \textsc{Title} & \textsc{Pr-\Faithful} & \textsc{Re-\Faithful} & \textsc{Pr-\Unfaithful} & \textsc{Re-\Unfaithful} \\\midrule
\multirow{7}{0.12\textwidth}{\gptturbo} & Only For The Week                  & 0.960                   & 0.972                & 0.333                    & 0.167                 \\
& Pet                                & 0.921                   & 0.923                & 0.333                    & 0.262                 \\
& She Is a Haunting                  & 0.957                   & 0.949                & 0.417                    & 0.333                 \\
& Six Scorched Roses                 & 0.794                   & 0.958                & 0.625                    & 0.288                 \\
& Sorrow and Bliss                   & 1.000                   & 0.868                & 0.139                    & 1.000                 \\
& Viciously Yours                    & 0.919                   & 0.919                 & 0.367                    & 0.354                 \\
& Yellowface                         & 0.952                   & 0.948                & 0.450                    & 0.438                 \\
\midrule
\multirow{7}{0.12\textwidth}{\claude} & Only For The Week            & 0.980                   & 0.971                & 0.571                    & 0.667                 \\
& Pet                          & 0.920                   & 0.910                & 0.429                    & 0.462                 \\
& She Is a Haunting            & 0.968                   & 0.968                & 0.571                    & 0.571                 \\
& Six Scorched Roses           & 0.919                   & 0.958                & 0.800                    & 0.667                 \\
& Sorrow and Bliss             & 1.000                   & 0. 966               & 0.429                    & 1.000                 \\
& Viciously Yours              & 0.931                   & 0.931                & 0.417                    & 0.417                 \\
& Yellowface                   & 0.963                   & 0.963                & 0.700                    & 0.700                 \\
\bottomrule
\end{tabular}
} % End scalebox
\caption{Average Precision (\textsc{Pr}) and Recall (\textsc{Re}) for the \textbf{Entire Book (EB)} approach (i.e., prompting the model with a claim and entire book as evidence) broken down by the rater models (\gptturbo\ and \claude), for each book.}
\label{tab:performance_metrics_WB}
\end{table}

\subsection{Prompting LLMs with the Entire Book (EB)}
\label{app:whole-book-approach}

Prompting LLMs with large chunks (\textit{entire books}) to evaluate the faithfulness of each claim is prohibitively expensive (see \S\ref{app:API_COST}). Hence, for this experiment, we select 7 books based on: (1) token length ($<$125K), and (2) presence of at least one \Unfaithful\ claim. 
This sub-dataset includes: (1) ``Yellowface,'' (2) ``Only For The Week,'' (3) ``Viciously Yours,'' (4) ``Six Scorched Roses,'' (5) ``Sorrow and Bliss,'' (6) ``She Is a Haunting,'' and (7) ``Pet.'' Table \autoref{tab:small_dataset_stat} shows the number of claims per label in the sub-dataset. Further details on each book can be found in \autoref{tab:books_in_corpus}.

\paragraph{Claim verification with the entire books} We prompt \claude\ and \gptturbo\ models with the entire book content and each claim in order to obtain the \Faithful/\Unfaithful\ labels. 

% \autoref{fig:confusion-nli} shows confusion matrix comparing human labels to models' predictions. 

\begin{figure}[htbp]
\centering
\includegraphics[width=1\linewidth]{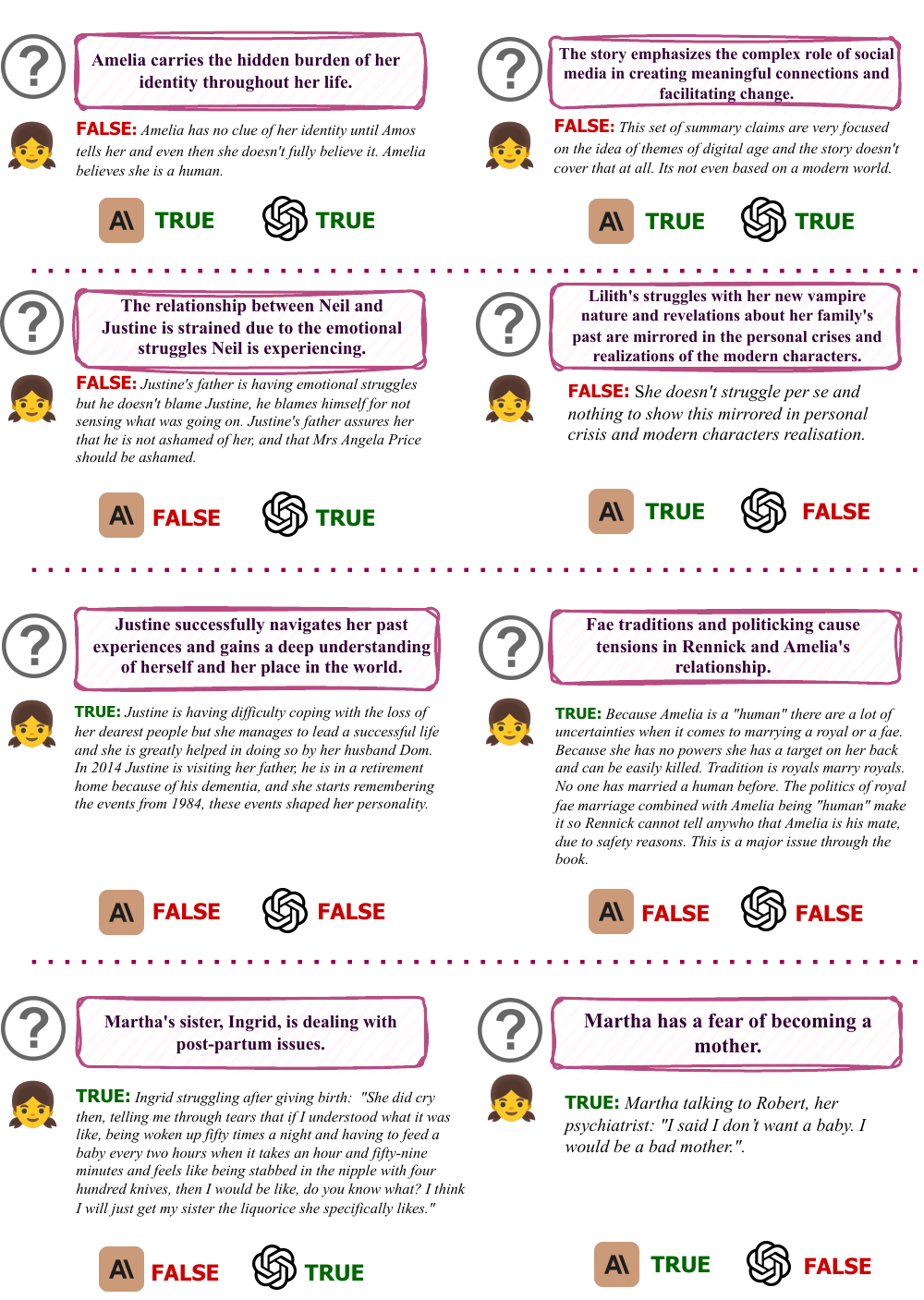}
\caption{Examples of claims accompanied by annotator labels and reasoning, along with predictions made by \claude\ and \gptturbo.}
\label{fig:examples-nli-results}
\end{figure}

\autoref{tab:gpt4_turbo_vs_claude3-nli-confusion} presents a confusion matrix broken down by claim source (i.e., the model that generated the claim) and prediction model (\claude\ and \gptturbo).
\autoref{fig:examples-nli-results} shows examples of misidentified labels by label-type and prediction model along with human labels and reasoning. \autoref{tab:performance_metrics_WB} shows average precision (\textsc{Pr}) and recall (\textsc{Re}) broken down by model and book.

% tab:small_dataset_stat
\begin{table}[htbp]
    \centering
    \scalebox{0.95}{
    \begin{tabular}{l|cccc}
    \toprule
    Model & Faithful & Unfaithful & Partial support & Can't verify \\
    \midrule
    \chatgpt & 111 & 22 & 20 & 8 \\
    \gpt & 152 & 11 & 30 & 3 \\
    \mixtral & 144 & 18 & 34 & 5 \\
    \gptturbo & 107 & 17 & 25 & 1 \\
   \claude & 140 & 1 & 17 & 0 \\
    \bottomrule
    \end{tabular}
    }
    \caption{Number of claims per label for each model in the sub-dataset of seven books.}
    \label{tab:small_dataset_stat}
\end{table}

% fig:confusion-nli}
% \begin{figure}[htbp]
% \centering
% \includegraphics[width=1\linewidth]{figures/confusion_nli_v8.pdf}
% \caption{Confusion matrix comparing human annotations to model predictions for the claim verification task. The left side shows the counts for predictions made by \claude\ contrasted with human labels. The right side shows the count for predictions made by \gptturbo\ contrasted with human labels. A more detailed breakdown is presented in \autoref{tab:gpt4_turbo_vs_claude3-nli-confusion}.}
% \label{fig:confusion-nli}
% \end{figure}

% fig:examples-nli-results

% tab:gpt4_turbo_vs_claude3-nli-confusion
\begin{table}[h]
\centering
\begin{tabular}{c|l|cc|cc}
\hline
\addlinespace
\multirow{2}{*}{\textbf{Human}} & \multirow{2}{*}{\textbf{Claim Source}} & \multicolumn{2}{c|}{\gptturbo} & \multicolumn{2}{c}{\claude} \\
& & \multicolumn{2}{c|}{\rule{3cm}{0.4pt}} & \multicolumn{2}{c}{\rule{3cm}{0.4pt}} \\
 & & {\color{red}\Unfaithful} & {\color{darkgreen}\Faithful} & {\color{red}\Unfaithful} & {\color{darkgreen}\Faithful} \\
 \addlinespace
\hline
\addlinespace
\multirow{5}{*}{\rotatebox{90}{{\color{red}\Unfaithful}}} 
 & \claude & 0 & 1 & 0 & 1 \\
 & \gptturbo & 3 & 14 & 9 & 8 \\
 & \gpt& 6 & 5 & 9 & 2 \\
 & \chatgpt & 8 & 14 & 13 & 9 \\
 & \mixtral & 11 & 7 & 10 & 8 \\
 \addlinespace
\hline
\addlinespace
\multirow{5}{*}{\rotatebox{90}{{\color{darkgreen}\Faithful}}} 
 & \claude & 10 & 130 & 7 & 133 \\
 & \gptturbo & 0 & 107 & 4 & 103 \\
 & \gpt & 20 & 132 & 10 & 142 \\
 & \chatgpt & 10 & 101 & 7 & 104 \\
 & \mixtral & 8 & 136 & 3z   & 141 \\
 \addlinespace
\hline
\end{tabular}
\caption{Count of labels predicted by \claude\ and \gptturbo\ contrasted with human-annotated labels, segmented by the model that generated each claim.}
\label{tab:gpt4_turbo_vs_claude3-nli-confusion}
\end{table}

\subsection{Ablation study}
\label{app:ablation_study}

\paragraph{Recall of the claim decomposition step}
\label{app:recall_decomp}
 we analyze the extracted claims on a subset of 20 summaries (371 sentences, 450 total extracted claims). We manually evaluate the quality of the extracted claims against the content of each summary. Calculating recall proved challenging due to the ambiguity in granularity (e.g., sentences, clauses, words). Notably, 3.8\% of the 371 sentences in the 20 summaries were omitted in the extracted claims. Of these omissions, 85.7\% were generic statements, and 14.3\% were minor details. Additionally, we observed a small percentage of omissions at the sub-sentential level (e.g., clauses), which did not impact the narrative. All These omissions can be broadly categorized into two types.
\begin{itemize}
    \item \textbf{Generic statements lacking substantive content:} For instance, “The narrative unfolds with intrigue, danger, and treacherous encounters” appears in the summary but is omitted in extracted claims. Note that this sentence only addresses things already covered by other extracted claims in a generic way, so omitting it has few consequences.
    \item \textbf{Insignificant details that contribute little to the narrative:} For instance, “Altha, a 17-century woman, stands trial unjustly accused of witchcraft due to her remarkable healing abilities which are misunderstood by her village” appears in the summary, but “misunderstood by her village” is omitted in the extracted claims. However, this is only a minor detail with little impact on the narrative.
\end{itemize}
Importantly, we confirmed that none of these discrepancies between the summaries and the extracted claims led to criticisms regarding omissions, chronological errors, or factual inaccuracies in the annotators' summary-level free-form comments.

\begin{figure}[htbp]
\centering
\includegraphics[width=0.8\linewidth]{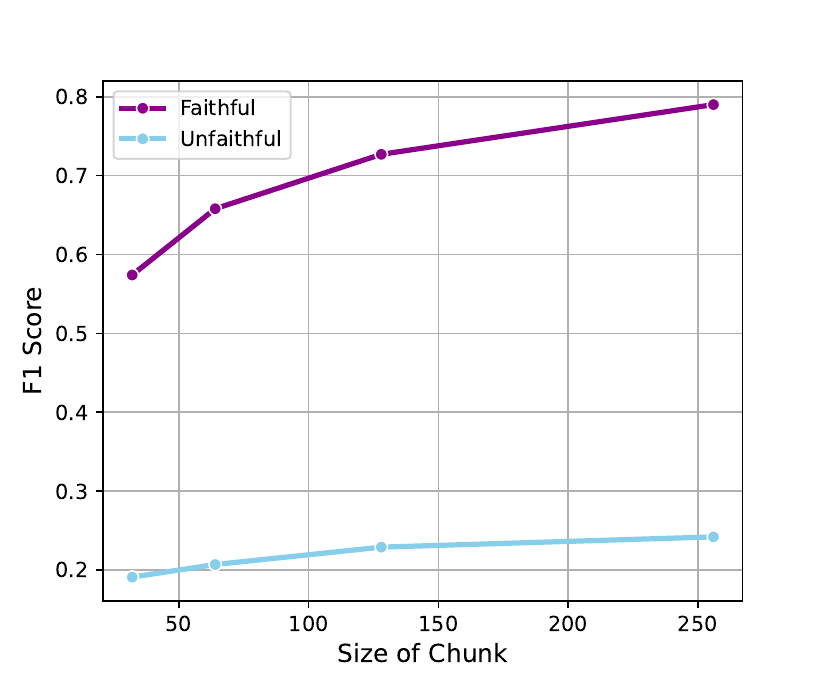}
\caption{F1 score varying chunk size for BM25.}
\label{fig:BM25_chunksize}
\end{figure}

\paragraph{Varying length of tokens used in BM25}
As we increase the length of BM25-retrieved passages, the overall performance improves (\autoref{fig:BM25_chunksize}). However, this approach remains less effective for identifying unfaithful claims than our best performing method, i.e., prompting the model with the content of the entire book. This is likely due to the fact that even longer passages may not provide the entire context needed for verification of broader claims.

\paragraph{Reasoning type of false positive cases}
We analyzed failure cases in which our auto-rater experiment, conducted on seven books using \claude\ and \gptturbo\, incorrectly marked an \Unfaithful\ claim as \Faithful. We annotated the types of reasoning required to verify these claims, as presented in \autoref{app:reasoning_type_false_positive}. The results indicate that approximately 75\% of these failure cases necessitate multi-hop reasoning across the book. This is significantly higher than the overall distribution of 62.8\% across the seven books, suggesting that our auto-raters struggle with multi-hop reasoning.

\begin{table}[h!]

\centering
\begin{tabular}{l c c c}
\toprule
\textbf{Reasoning Type} & \textbf{\claude} & \textbf{\gptturbo} & \textbf{Both Models} \\
                       & \textbf{(28 examples)}            & \textbf{(37 examples)}               & \textbf{(24 examples)}                          \\
\midrule
Indirect               & 75.0\%                            & 73.0\%                               & 75.0\%                                          \\
Direct                 & 14.3\%                            & 10.8\%                               & 12.5\%                                          \\
Subjective             & 7.1\%                             & 10.8\%                               & 8.3\%                                           \\
Extra info             & 3.6\%                             & 5.4\%                                & 4.2\%                                           \\
\bottomrule
\end{tabular}
\caption{Reasoning type distribution for false positives case by each model}
\label{app:reasoning_type_false_positive}
\end{table}

% fig:BM25_chunksize

\subsection{API costs}
\label{app:API_COST}

\paragraph{Generating book-length summaries} The total cost of summarization for all 130 summaries amounted to about \$288 USD.\footnote{\$64.6 for \claude, \$169.4 for \gpt, \$47.5 for \gptturbo, \$2.8 for \chatgpt, and \$3.4 for \mixtral. All in USD.}

\paragraph{Extracting claims} The total cost of claim extraction for all 130 summaries amounted to about \$8 USD, as the input and output sequence is relatively short.

\paragraph{Prompting with the entire book} This experiment costed roughly \$720 USD for \gptturbo\ and \$1070 USD for \claude\  (corresponding to the last two columns in \autoref{tab:benchmark_results_hay}).

\end{document}